\documentclass[journal]{IEEEtran}
\usepackage{amsmath,amsfonts}
\usepackage{array}
\usepackage{hyperref}
\usepackage{stfloats}
\usepackage{url}
\usepackage{verbatim}
\usepackage{graphicx}
\usepackage{cite}
\usepackage{graphicx}
\usepackage{url}
\usepackage{cleveref}
\usepackage{booktabs}
\usepackage{algorithm}
\usepackage{algpseudocode}
\usepackage{multirow}
\usepackage{bm}
\usepackage{colortbl}
\usepackage[table]{xcolor}
\usepackage{subfigure}
\usepackage{lipsum} 
\usepackage{amssymb}

\usepackage{xspace}
\usepackage{pifont}
\usepackage{siunitx}
\newcommand{\eg}{\textit{e.g.}\@\xspace}
\newcommand{\ie}{\textit{i.e.}\@\xspace}

\begin{document}

\title{Efficient Diffusion-Based 3D Human Pose Estimation with Hierarchical Temporal Pruning}

\author{ 
    Yuquan Bi, Hongsong Wang, Xinli Shi,~\IEEEmembership{Senior Member,~IEEE,} Zhipeng Gui,~\IEEEmembership{Member,~IEEE,} Jie Gui,~\IEEEmembership{Senior Member,~IEEE,} and Yuan Yan Tang,~\IEEEmembership{Life Fellow,~IEEE}

\thanks{\textit{Corresponding author: Jie Gui and Hongsong Wang.}}
\thanks{Yuquan Bi is with the School of Cyber Science and Engineering, Southeast University, Nanjing 211189, China (e-mail: yuquanbi@seu.edu.cn).}
\thanks{H. Wang is with School of Computer Science and Engineering, Key Laboratory of New Generation Artificial Intelligence Technology and Its Interdisciplinary Applications, Ministry of Education, Southeast University, Nanjing 210096, China (e-mail: hongsongwang@seu.edu.cn).}
\thanks{Xinli Shi is with the National Center for Applied Mathematics, Southeast University, Nanjing 211189, China (e-mail: xinli shi@seu.edu.cn).}
\thanks{Z. Gui is with the School of Remote Sensing and Information Engineering and the Collaborative Innovation Center of Geospatial Technology,
Wuhan University, Wuhan 430079, China (e-mail: zhipeng.gui@whu.edu.cn).}
\thanks{Jie Gui is with the School of Cyber Science and Engineering, Southeast University, Nanjing 211189, China, also with Purple Mountain Laboratories, Nanjing 211111, China, and also with the Engineering Research Center of
Blockchain Application, Supervision and Management (Southeast University), Ministry of Education, Nanjing 210000, China (e-mail: guijie@seu.edu.cn).}
\thanks{Yuan Yan Tang is with the Department of Computer and Information
Science, University of Macau, Macau, China, and also with Faculty of Science and Technology, UOW College Hong Kong, Hong Kong, China (e-mail: yytang@um.edu.mo).}
}

\markboth{Manuscript for IEEE Transactions on Circuits and Systems for Video Technology}%
{Shell \MakeLowercase{\textit{et al.}}: A Sample Article Using IEEEtran.cls for IEEE Journals}


\maketitle

\begin{abstract}

Diffusion models have demonstrated strong capabilities in generating high-fidelity 3D human poses, yet their iterative nature and multi-hypothesis requirements incur substantial computational cost. In this paper, we propose an efficient diffusion-based 3D human pose estimation framework with a Hierarchical Temporal Pruning (HTP) strategy, which dynamically prunes redundant pose tokens across both frame and semantic levels while preserving critical motion dynamics. HTP operates in a staged, top-down manner: (1) Temporal Correlation-Enhanced Pruning (TCEP) identifies essential frames by analyzing inter-frame motion correlations through adaptive temporal graph construction; (2) Sparse-Focused Temporal MHSA (SFT MHSA) leverages the resulting frame-level sparsity to reduce attention computation, focusing on motion-relevant tokens; and (3) Mask-Guided Pose Token Pruner (MGPTP) performs fine-grained semantic pruning via clustering, retaining only the most informative pose tokens. Experiments on Human3.6M and MPI-INF-3DHP show that HTP reduces training MACs by 38.5\%, inference MACs by 56.8\%, and improves inference speed by an average of 81.1\% compared to prior diffusion-based methods, while achieving state-of-the-art performance.
\end{abstract}


\section{Introduction}
\label{sec:intro}

\IEEEPARstart
3D human pose estimation (HPE) from monocular videos is a fundamental task advancing rapidly in recent years for its significant applications in action recognition \cite{intro-action1, intro-action2, intro2, intro-action3, intro-action4}, human-robot interaction \cite{intro3, intro4,intro-robot1}, and virtual reality \cite{intro5, intro6, wang2021mars}. Benefiting from the excellent performance of 2D pose detectors \cite{cpn, sh, hrnet,intro-2d1,intro-2d2,intro-2d3}, the 2D-to-3D lifting pipeline \cite{2d-to-3d01,2d-to-3d2,2d-to-3d3,2d-to-3d4,due,intro-2dto3d1,intro-2dto3d2,gkonet} has become dominant due to its high precision and lightweight nature.

\begin{figure}[t]
\centering
\includegraphics[width=1\columnwidth]{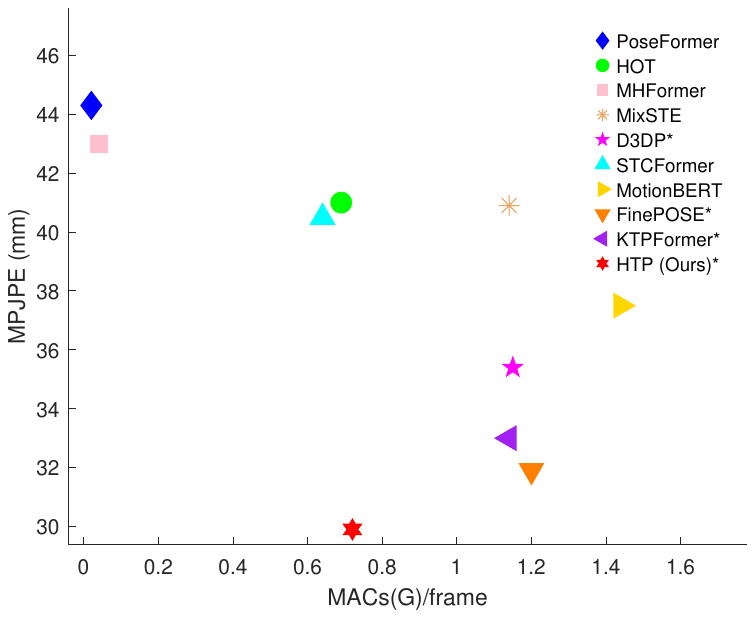}
\caption{\textbf{MACs and MPJPE of different methods on the Human3.6M dataset.} We achieve the best performance while demonstrating highly competitive MACs results. $^{\ast}$ indicates diffusion-based methods.}
\label{fig_intro}
\end{figure}

The 2D-to-3D lifting pipeline lacks depth priors and suffers from ambiguity. To mitigate this issue, recent works incorporate temporal correlations across video frames into the pose reconstruction process. For example, many transformer-based architectures \cite{hdformer, poseformer, poseformerv2, stcformer, motionbert, agformer, mixste, p-stmo, tc-mixste, fpm, dualformer} effectively capture long-range temporal dependencies by encoding the joint-level semantics of each video frame into pose tokens, achieving promising performance even on extremely long video sequences. However, the computational cost for spatial-temporal modeling in self-attention (SA) increases quadratically as the number of frames increases, resulting in substantial computational overhead.

Diffusion-based 3D HPE leverages transformer architectures to resolve depth ambiguity through the iterative refinement of high-fidelity 3D pose generation. These methods employ a transformer-based diffusion process, which requires $K$ steps of iterative refinement to generate $H$ pose hypotheses during inference. However, the inherent computational complexity of diffusion models, combined with the transformer-based SA mechanisms, leads to significant resource demands. 
For instance, processing a 243-frame video sequence with D3DP \cite{d3dp} requires 1.15G MACs per frame during training, but this increases to 228.8G per frame during inference with $H=20$ and $K=10$. Although adjusting $H$ and $K$ offers a theoretical trade-off between accuracy and efficiency, the combined cost of diffusion steps and transformer operations makes it challenging to achieve both simultaneously. 

A straightforward approach to reduce the primary computational bottleneck in 3D HPE is to eliminate redundant pose tokens in the temporal SA calculations. Existing methods typically adopt two disjoint strategies: (1) Frame-level pruning that discards adjacent frames under static redundancy assumptions, and (2) Semantic-level sparsification that clusters low-information tokens via heuristic criteria. While these single-stage strategies effectively reduce computation, they often overlook subtle yet crucial motion transitions. More importantly, such approaches are not well-suited for diffusion-based 3D HPE, where pose reconstruction unfolds iteratively across multiple noise levels. Simply applying naive pruning risks discarding informative content at intermediate steps, thereby compromising motion continuity and stability. To address these challenges, we propose an efficient diffusion-based 3D human pose estimation framework with Hierarchical Temporal Pruning (HTP), which operates across both frame-level and semantic-level stages to preserve essential motion dynamics throughout the denoising process. By selectively retaining key frames and salient pose tokens at each denoising iteration, HTP maintains the integrity of global motion patterns while reducing computational cost.

Specifically, we implement HTP through a structured, hierarchical pruning framework across both frame and semantic levels. 
First, the Temporal Correlation-Enhanced Pruning (TCEP) module analyzes temporal correlations across video frames. Each node represents a video frame, and we compute a dense correlation matrix to measure inter-frame similarity. Based on this, our Correlation-Enhanced Node Selection Algorithm constructs a dynamic temporal graph and selects a subset of nodes with strong temporal relevance as representative frames. A Sparse Binary Mask $\mathbf{M}$ is generated to store the retained temporal relationships. 
Second, based on the temporal correlations identified by TCEP, the Sparse-Focused Temporal Multi-Head Self-Attention (SFT MHSA) uses $\mathbf{M}$ to guide attention toward motion-relevant pose tokens. By restricting attention to key frames, SFT MHSA reduces computational overhead while preserving the model’s ability to capture global temporal dependencies. 
Finally, we apply the Mask-Guided Pose Token Pruner (MGPTP), which integrates frame-level correlations from TCEP and sparse pose token representations from SFT MHSA. MGPTP discards redundant pose tokens while preserving tokens critical to motion fidelity using a density-aware strategy guided by the sparse mask $\mathbf{M}$. 
Together, these modules form a cohesive hierarchical denoising framework that enhances computational efficiency and preserves motion fidelity in diffusion-based 3D HPE. As shown in Fig. \ref{fig_intro}, HTP reduces training MACs by an average of $38.5\%$, setting a new standard in efficient diffusion-based 3D HPE. The contributions of this paper are as follows:
\begin{itemize}
\item We propose Hierarchical Temporal Pruning (HTP), a unified hierarchical pruning framework integrated into diffusion-based 3D HPE that reduces both frame- and token-level redundancy to improve efficiency, overcoming the limitations of previous single-stage strategies.
\item TCEP, SFT MHSA, and MGPTP operate under a unified sparse constraint $\mathbf{M}$ to collaboratively reduce temporal redundancy and preserve motion-critical dynamics. All modules are plug-and-play and compatible with both diffusion- and transformer-based 3D HPE pipelines.
\item Extensive experiments on Human3.6M and MPI-INF-3DHP show that HTP achieves state-of-the-art accuracy while reducing training MACs by 38.5\%, inference MACs by 56.8\%, and boosting FPS by 81.1\% on average.
\end{itemize}

\section{Related Work}
\label{sec:related_work}

\subsection{Transformer-Based 3D HPE} 
The Transformer, first proposed by Vaswani et al. \cite{transformer1}, has demonstrated remarkable performance in computer vision (CV) tasks \cite{transformer2,vit,wang2020deep,transformer3, transformer4}, as the self-attention mechanism has a strong ability to capture long-range dependencies. This characteristic makes it particularly well-suited for 3D HPE. PoseFormer \cite{poseformer} was the first to adopt the vision transformer as a backbone network for video-based 3D HPE. MixSTE \cite{mixste} alternates between spatial and temporal transformer blocks to capture spatio-temporal features, providing 3D pose estimates for each frame in the input sequence. DualFormer \cite{dualformer} further enhances performance via dual-path attention across joints and frames. GKONet \cite{gkonet} incorporates geometric priors into a graph-guided transformer for structure-aware prediction. STCFormer \cite{stcformer} reduces computational complexity by separately modeling spatial and temporal components of input joints. MotionBERT \cite{motionbert} introduces a dual-stream spatial-temporal transformer to model long-range spatial-temporal relationships, and is finetuned for skeletal joint-based tasks. However, the quadratic complexity of self-attention in spatial-temporal modeling results in substantial computational overhead.

\subsection{Diffusion-Based 3D HPE} 
Diffusion models are a class of generative models that progressively degrade observed data by adding noise, and then restore the original data through a reverse denoising process. These models have demonstrated promising results across various applications, such as image$/$video generation \cite{image_gen1,image_gen2,wang2025noisy}, super-resolution \cite{surper1,image_gen2}, and Human Motion Generation \cite{motion_gen1,motion_gen2,motion_gen3,motion_gen4}. Recently, several 3D HPE methods based on diffusion models \cite{diffpose, d3dp, d3dp-jour, finepose, ktpformer} have been proposed to generate high-fidelity 3D human poses, aiming to address the challenge of intrinsic depth ambiguity. D3DP \cite{d3dp, d3dp-jour} integrates a denoiser based on MixSTE \cite{mixste} to reconstruct noisy 3D poses by assembling joint-by-joint multiple hypotheses. FinePose \cite{finepose} learns modifiers for different human body parts to describe human movements at multiple levels of granularity. KTPFormer \cite{ktpformer} incorporates the anatomical structure of the human body and joint motion trajectories across frames as prior knowledge to learn spatial and temporal correlations. However, the reliance on iterative refinement ($K$ steps) and multiple hypotheses ($H$ samples) significantly increases the computational burden, posing greater efficiency challenges compared to transformer-based counterparts. Distinct from prior diffusion-based approaches \cite{d3dp, finepose, ktpformer} that primarily focus on enhancing generation fidelity, HTP is explicitly designed to mitigate this computational bottleneck. By introducing a hierarchical temporal pruning strategy, it effectively optimizes the trade-off between efficiency and performance.

\begin{figure*}[t]\centering
\includegraphics[width=1 \textwidth]{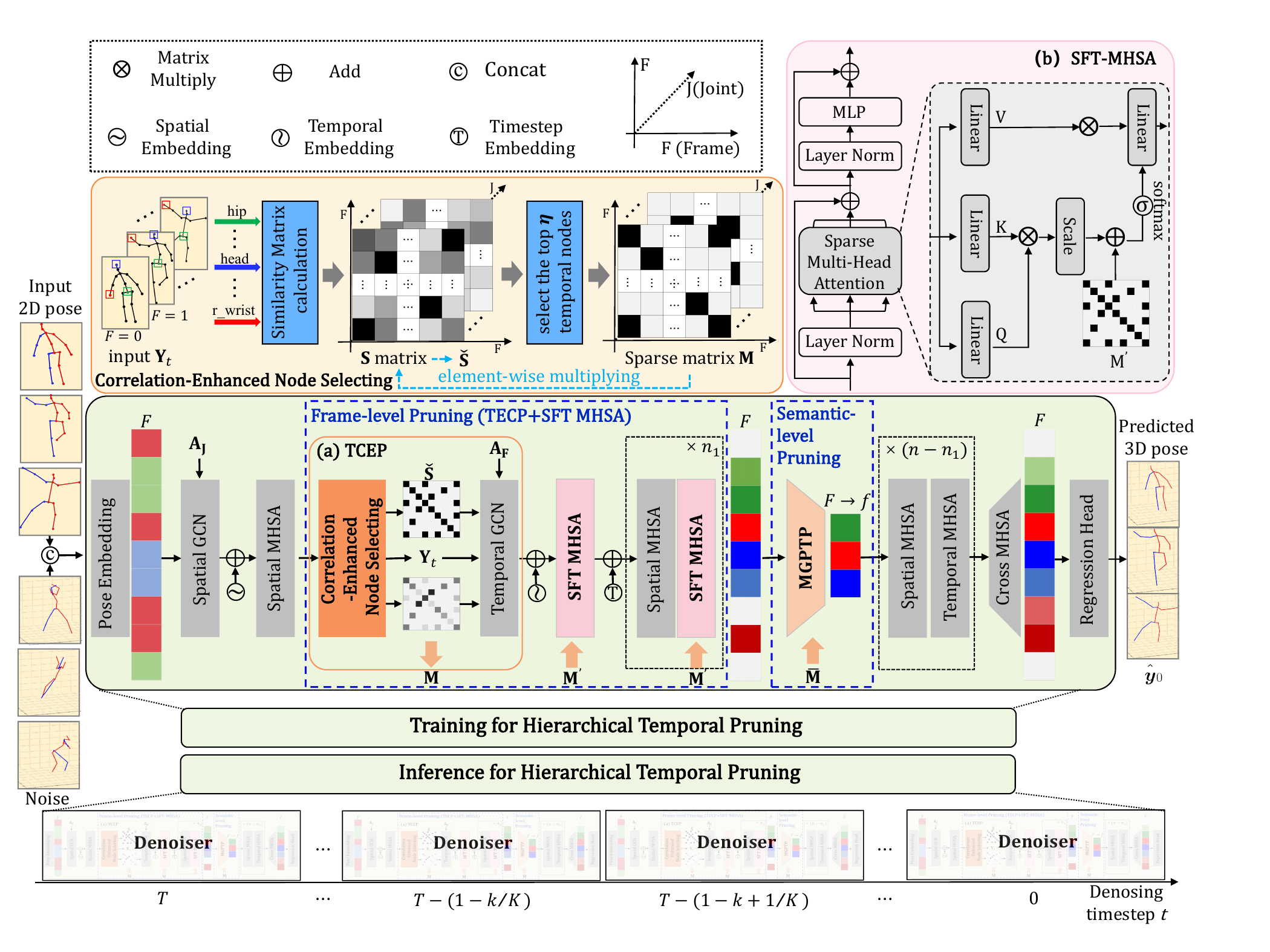}
\caption{\label{fig_architecture} \textbf{The architecture of the proposed HTP.} The framework is structured into two hierarchical pruning phases: \textbf{(a) Frame-level Pruning} and \textbf{(b) Semantic-level Pruning}. At each diffusion step, the input $[\boldsymbol{y}_t\oplus\boldsymbol{x}]$ is embedded and processed by the Spatial GCN \cite{agformer} and Spatial MHSA. \textbf{In Phase (a)}, TCEP first infers a sparse temporal mask $\mathbf{M}$ and pruned features, which then guide the SFT MHSA in modeling frame-to-frame dependencies on the full sequence length $F$ with reduced redundancy. \textbf{In Phase (b)}, MGPTP physically condenses the sequence from $F$ to $f$ by aggregating representative tokens. Finally, Cross MHSA restores the original length $F$ for prediction. Note that $\mathbf{M}'$ and $\overline{\mathbf{M}}$ denote variants of $\mathbf{M}$ adapted for SFT MHSA and MGPTP, respectively. Pose embedding, Spatial MHSA, and full Temporal MHSA are standard operations in 3D HPE.} 
\end{figure*}

\subsection{Improving Efficiency in 3D HPE} 
Enhancing computational efficiency in 3D human pose estimation is crucial for real-world applications, especially in resource-limited environments. While prior works have explored pruning strategies to improve temporal modeling efficiency, most adopt a single-level approach. Frame-level methods, such as DeciWatch \cite{deciwatch}, G-SFormer \cite{gsformer}, and Uplift \cite{uplift}, reduce computational cost by sampling or scheduling representative frames based on temporal redundancy. Token-level methods, including HOT \cite{hot} and P-STMO \cite{p-stmo}, focus on semantic sparsification by removing low-saliency tokens or clustering similar features. While these single-strategy approaches help reduce complexity, they often fail to account for interactions between temporal structure and semantic content. This can lead to suboptimal retention of motion-critical information, particularly in dynamic scenes or under iterative refinement processes like diffusion. In contrast, our approach integrates pruning at both frame and semantic levels within a unified framework. This hierarchical design enables more informed token selection and better preserves motion coherence throughout the denoising, providing a more robust and adaptive solution than methods relying on a single strategy.

\section{Preliminary of Diffusion-Based 3D HPE}
\label{sec:preliminary}

Diffusion models are generative frameworks that characterize data distributions via a time-dependent diffusion process with two phases: $\text{(1)}$ \emph{a training phase}, where data is progressively perturbed by adding noise, and a denoiser is trained to reverse this perturbation, and $\text{(2)}$ \emph{an inference phase}, where the trained denoiser reconstructs the original, uncorrupted data.

\subsection{Training}
Starting from the ground truth input 3D pose $\boldsymbol{y}_0$, a sequence of noisy samples $\{\boldsymbol{y}_t\}_{t=1}^{T}$ is generated, where $T$ denotes the total number of timesteps. During this process, standard Gaussian noise is introduced to $\boldsymbol{y}_0$, progressively transforming it into a Gaussian distribution $\boldsymbol{y}_T \sim p_T$. Following DDPMs \cite{ddpm}, the perturbation of $\boldsymbol{y}_t$ can be expressed as:

\begin{equation}
q(\boldsymbol{y}_t \mid \boldsymbol{y}_0) := \sqrt{\overline{\alpha}_t} \boldsymbol{y}_{0} + \sqrt{1 - \overline{\alpha}_t} \epsilon,
\end{equation}
with $\alpha_t = 1 - \beta_t$, $\overline{\alpha}_t = \prod_{s=1}^{t} \alpha_s$, and $\epsilon\sim\mathcal{N}(0,\boldsymbol{I})$. where $\{\beta_t\}_{t=1}^{T}$ denotes the variance schedule.

Subsequently, $\boldsymbol{y}_t$ is passed to the Denoiser conditioned on 2D keypoints $\boldsymbol{x}$ and timestep $t$ to reconstruct the original 3D pose $\boldsymbol{\hat{y}}_{0}$. The entire framework is optimized through 
the standard MPJPE (Mean Per Joint Position Error) loss, which minimizes the average Euclidean distance between the predicted and ground-truth 3D joint positions:

\begin{equation}
\mathcal{L}=\frac{1}{J}\sum_{k=1}^J\|\boldsymbol{y}_{0}-\boldsymbol{\hat{y}}_{0}\|_2.
\end{equation}

\subsection{Inference}
During inference, the reverse diffusion process is applied to recover the original 3D pose $\boldsymbol{y}_0$ by iteratively denoising the noisy sample $\boldsymbol{y}_T \sim p_T$. Following D3DP \cite{d3dp}, we adopt the multi-hypothesis strategy within the DDIM \cite{ddim} framework, focusing on predicting the original input rather than the noise. The reverse process can be articulated as follows:

\begin{equation}
\boldsymbol{y}_{h,t-1} = \sqrt{\overline{\alpha}_{t-1}} \cdot\boldsymbol{\hat{y}}_{h,0} + \sqrt{1 - \overline{\alpha}_{t-1} - \sigma_{t}^2} \hat{\epsilon}_{t} + \sigma_{t} \epsilon,
\end{equation}
where $t$ and $t-1$ are the adjacent timesteps in the subset $\tau \subset \{1, \ldots, T\}$. $h \subset \{0, \ldots, H\}$, $\epsilon\sim\mathcal{N}(0,\boldsymbol{I})$, and 

\begin{equation}
\begin{split}
    \hat{\epsilon}_{t} &= \left(\boldsymbol{y}_{h,t}-\sqrt{\overline{\alpha}_{t}}\cdot\boldsymbol{\hat{y}}_{h,0}\right)/\sqrt{1-\overline{\alpha}_{t}} \\
    \sigma_{t} &= \sqrt{(1-\overline{\alpha}_{t-1})/(1-\overline{\alpha}_{t})}\cdot\sqrt{1-\overline{\alpha}_{t}/\overline{\alpha}_{t-1}}. 
\end{split}
\end{equation}

Beginning by sampling $H$ initial 3D poses $\boldsymbol{y}_{h,T}$ from a unit Gaussian and then fedding the samples into the denoiser to produce $H$ viable 3D pose hypotheses $\boldsymbol{y}_{h,0}$. This process is repeated iteratively for $K$ steps, with the timestep $t$ updated as $T\left(1 - \frac{k}{K}\right)$ at each iteration $k \in [1, K]$.

In line with other diffusion-based methods \cite{d3dp, finepose, ktpformer}, we employ the Joint-Wise Reprojection-Based Multi-Hypothesis Aggregation (JPMA) technique \cite{d3dp} to aggregate and evaluate the final 3D pose predictions.

\section{Method}
\label{sec:method}


\begin{table*}[ht]
    \centering
    \caption{\textbf{Quantitative Comparison with the SOTA Methods on the Human3.6M Dataset.} $F$: The Number of Input Frames. CE: Estimating Center Frame Only. Detector: Using CPN~\cite{cpn} and SH~\cite{sh} as the 2D Keypoints Detector to Generate the Inputs, or Using the Ground Truth 2D Keypoints as Inputs. $^{\dagger}$ Indicates the Scratch Setting in~\cite{motionbert}, and $^{\ddagger}$ Indicates the Finetune Setting in~\cite{motionbert}. HTP (Ours) Utilizes the Default D3DP Backbone; Variants like HTP w/ MixSTE Demonstrate Plug-and-play Generalization. The Best and Second-Best Results Are Highlighted in \textbf{Bold} and \underline{Underline} Formats.}
    \renewcommand{\arraystretch}{1.2}
    \resizebox{1.0\textwidth}{!}
    {
    \begin{tabular}{l c  | c  c c ccc cc c c}
        \toprule
        \multirow[c]{2}{*}{Method} & \multirow[c]{2}{*}{Type}  & \multirow[c]{2}{*}{Publication} & \multirow[c]{2}{*}{$F$} & \multirow[c]{2}{*}{CE} & \multicolumn{3}{c}{Human3.6M (DET)} & \multicolumn{2}{c}{Human3.6M (GT)} & \multirow[c]{2}{*}{MACs (G)} & \multirow[c]{2}{*}{Params (M)} \\
        \cmidrule(lr){6-8} \cmidrule(lr){9-10}
        & & & & & Detector & MPJPE $\downarrow$ & P-MPJPE $\downarrow$ & Detector & MPJPE $\downarrow$ & \\
        \midrule
        TCN~\cite{tcn} & CNN & CVPR'19 & 243 & -- & CPN & 46.8 & 36.5 & GT & 37.8 & -- & -- \\
        GLA-GCN~\cite{gla-gcn} & GCN & ICCV'23 & 243 & \checkmark & CPN & 44.4 & 34.8 & GT & 21.0  & -- & --  \\
        FTCM~\cite{ftcm} & MLP & TCSVT'24 & 351 & \checkmark & CPN & 45.3 & 35.3 & GT & 28.2 & 1.5 & 4.7 \\
        PoseMamba-X~\cite{posemamba} & Mamba & AAAI'25 & 243 & \ding{53} & CPN & 37.1 & 31.5 & GT & -- & 109.9 & 26.5 \\
        SAMA-L~\cite{sama} & Mamba & ICCV'25 & 243 & \ding{53} & CPN & 36.9 & 31.3 & GT & -- & 53.2 & 17.3 \\
        \midrule
        PoseFormer~\cite{poseformer} & Transformer & ICCV'21 & 81 & \checkmark & CPN & 44.3 & 36.5  & GT & 31.3 & 1.6 & 9.6  \\
        P-STMO~\cite{p-stmo} & Transformer & ECCV'22  & 243 & \checkmark & CPN   & 42.8 & 34.4 & GT & 29.3 & 1.7 & 7.0   \\
        PoseFormerV2~\cite{poseformerv2} & Transformer & CVPR'23 & 243 &\checkmark & CPN & 45.2 & 35.6 & GT & 35.5 & 2.1 & 14.4  \\
        MHFormer~\cite{mhformer} & Transformer & CVPR'22 & 351 & \checkmark & CPN & 43.0 & 34.4 & GT & 30.5  & 9.6 & 24.7   \\
        MixSTE~\cite{mixste} & Transformer & CVPR'22  & 243 & \ding{53} & CPN & 40.9 & 30.6 & GT & 21.6  & 139.0  & 33.8    \\
        STCFormer~\cite{stcformer} & Transformer & CVPR'23 & 243 & \ding{53} & CPN & 40.5 & 31.8  & GT & 21.3 & 78.2 & 18.9  \\
        MotionBERT~\cite{motionbert}$^{\dagger}$ & Transformer & ICCV'23 & 243 & \ding{53} & SH & 39.2 & -- & GT & 17.8  & 174.8 & 42.5   \\
        MotionBERT~\cite{motionbert}$^{\ddagger}$ & Transformer & ICCV'23 & 243 & \ding{53} & SH & 37.5 & -- & GT & 16.9  & 174.8 & 42.5   \\
        HOT~\cite{hot} & Transformer & CVPR'24 & 243 & \ding{53} & CPN & 41.0 & -- & GT & -- & 83.8 & 35.0  \\
        TC-MixSTE~\cite{tc-mixste} & Transformer & TMM'24 & 243 & \ding{53} & CPN & 39.9 & 31.9 & GT & 21.1 & -- & -- \\
        DualFormer~\cite{dualformer} & Transformer & TCSVT'24 & 351 & \ding{53} & CPN & 42.8 & 34.4 & GT & 28.9 & -- & -- \\
        \rowcolor{black!10} HTP w/ MixSTE & Transformer  & -- & 243 & \ding{53} & CPN & 39.9 & 29.9 & GT & 20.7  & 87.6 & 36.4 \\
        \rowcolor{black!10} HTP w/ MotionBERT$^{\dagger}$ & Transformer  & -- & 243 & \ding{53} & SH & 38.9 & -- & GT & 17.7  & 101.7 & 47.6 \\
        \midrule
        Diffpose~\cite{diffpose} & Diffusion & CVPR'23 & 243 & \ding{53} & CPN & 36.9 & 28.7 & GT & 18.9  & -- & --  \\
        Diffpose~\cite{diffpose2} & Diffusion & ICCV'23 & 64 & \ding{53} & -- & 42.9 & 30.8 & GT & --  & -- & --  \\
        D3DP~\cite{d3dp} & Diffusion & ICCV'23 & 243 &\ding{53} & CPN & 35.4 & 28.7 & GT & 18.4 & 139.1 & 34.8  \\
        FinePOSE~\cite{finepose} & Diffusion & CVPR'24 & 243 & \ding{53} & CPN & \underline{31.9} & \underline{25.0} & GT & \textbf{16.7}  & 146 & 200.6  \\
        KTPFormer~\cite{ktpformer} & Diffusion & CVPR'24 & 243 & \ding{53} & CPN & 33.0 & 26.2 & GT & 18.1 & 139.1 & 36.3  \\
        \rowcolor{black!10} HTP (Ours)  & Diffusion &  -- & 243 & \ding{53} & CPN & \textbf{29.9} & \textbf{23.3} & GT & \textbf{16.7}  & 87.7 & 37.5 \\
        \bottomrule
    \end{tabular}
    }
    \label{tab:h36m_res1}
\end{table*}

\begin{algorithm}[t]
\small
\caption{Generation and Utilization of $\mathbf{M}$}
\label{alg:tcep_mask}
\begin{algorithmic}[1]

\Require 
    Input Pose Tokens $\mathbf{Y} \in \mathbb{R}^{J \times F \times D}$, 
    Neighbor Count $\eta$
\Ensure 
    $\mathbf{M} \in \mathbb{R}^{J \times F \times F}$, $\mathbf{M}' \in \mathbb{R}^{J \times F \times F}$, $\overline{\mathbf{M}} \in \mathbb{R}^{F \times F}$

\State \textcolor{gray}{\textbf{Phase 1: Joint-wise Mask Generation (Section \ref{tcep})}}
\State \textbf{Initialize} $\mathbf{M} \leftarrow \{0\}^{J \times F \times F}$

\For{each joint index $j \in \{1, \dots, J\}$} 
    \State $\boldsymbol{Y}_t^{(j)} \leftarrow \boldsymbol{Y}_t [j, :, :] \in \mathbb{R}^{F \times D}$ \Comment{Extract temporal features}
    
    \State \textcolor{gray}{// 1.1 Compute Pairwise Similarity (Eq. \ref{eq6})}
    \State $\mathbf{S}^{(j)} \leftarrow (\boldsymbol{Y}_t^{(j)} \cdot (\boldsymbol{Y}_t^{(j)})^\top) / \sqrt{D}$ 
    
    \State \textcolor{gray}{// 1.2 Exclude self-loops and select Top-$\eta$ Neighbors, }
    \State $\mathbf{S}^{(j)}_{\text{masked}} \leftarrow \text{MaskDiagonal}(\mathbf{S}^{(j)}, -\infty)$
    \State $\mathcal{I}_{\text{top}} \leftarrow \text{TopK}(\mathbf{S}^{(j)}_{\text{masked}}, k=\eta, \text{dim}=-1).\text{indices}$ 
    \State \textcolor{gray}{// $\mathcal{I}_{\text{top}}[p]$ contains the set of neighbor indices $\{q\}$ for $p$}
    
    \State \textcolor{gray}{// 1.3 Construct Symmetric Connectivity (Eq. \ref{eq7})}
    \For{each frame $p \in \{1, \dots, F\}$}
        \State $\mathbf{M}[j, p, p] \leftarrow 1$ \Comment{Self-loop ($p=q$)}
        \State $\mathcal{Q}_p \leftarrow \mathcal{I}_{\text{top}}[p]$ \Comment{Get neighbor indices $q$ for frame $p$}
        \State $\mathbf{M}[j, p, \mathcal{Q}_p] \leftarrow 1$ \Comment{Forward: $q \in \text{Top}_{p}(\mathbf{S}^{(j)}, \eta)$}
        \State $\mathbf{M}[j, \mathcal{Q}_p, p] \leftarrow 1$ \Comment{Symmetric: $p \in \text{Top}_{q}(\mathbf{S}^{(j)}, \eta)$}
    \EndFor
\EndFor
\State \textcolor{gray}{// $\mathbf{M}$ is now the stacked binary mask $\in \{0, 1\}^{J \times F \times F}$}

\Statex \textcolor{gray}{\textbf{Phase 2: Application in SFT MHSA (Section \ref{sft-mhsa})}}
\State \textcolor{gray}{// Convert binary mask to additive attention bias (Eq. \ref{eq11})}
\State $\mathbf{M}' \leftarrow \text{Where}(\mathbf{M} == 1, \ 0, \ -\infty)$ 

\Statex \textcolor{gray}{\textbf{Phase 3: Application in MGPTP (Section \ref{mgptp})}}
\State \textcolor{gray}{// Aggregate across joints}
\State $\overline{\mathbf{M}} \leftarrow \text{Mean}(\mathbf{M}, \text{dim}=0) \in \mathbb{R}^{F \times F}$ 

\end{algorithmic}
\end{algorithm}

\subsection{Overview}

Following the formulation in Sec.~\ref{sec:preliminary}, our framework estimates the clean 3D pose $\boldsymbol{y}_{0}$ from its noisy observation $\boldsymbol{y}_{t}$ conditioned on 2D keypoints $\boldsymbol{x}$. As illustrated in Fig.~\ref{fig_architecture}, the input pair is first projected into a high-dimensional representation and processed by the Spatial GCN \cite{agformer} and the Spatial MHSA to encode skeletal topology. The output $\boldsymbol{Y}_{t}$ then undergoes our core Hierarchical Temporal Pruning (HTP) strategy, which operates in a coarse-to-fine manner:

\textbf{Frame-Level Pruning:} This phase focuses on filtering redundancy while maintaining full temporal resolution ($F$). First, the \textit{Temporal Correlation-Enhanced Pruning (TCEP)} module initiates the hierarchy by establishing a sparse topology to filter static frames. Subsequently, the diffusion timestep embedding $\mathcal{F}(t)$ is injected into the feature map. The \textit{Sparse-Focused Temporal MHSA (SFT MHSA)} performs transitional refinement: it models long-range dependencies strictly within this sparse structure. By enhancing the feature discriminability of the retained frames, it acts as a semantic bridge, preparing the representation for the subsequent hard pruning.

\textbf{Semantic-Level Pruning:} Advancing the hierarchy, the \textit{Mask-Guided Pose Token Pruner (MGPTP)} executes "hard-pruning" by physically compressing the sequence length from $F$ to $f$. It aggregates the refined tokens from SFT MHSA into high-level descriptors, achieving deep semantic abstraction.

The condensed sequence $\bar{\boldsymbol{Y}}_t$ is then processed by $n-n_1$ standard encoder blocks for deep refinement. Finally, a Cross MHSA restores the full temporal resolution (from $f$ back to $F$) for the final prediction $\hat{\boldsymbol{y}}_0$. During inference, we follow the reverse process detailed in Sec.~\ref{sec:preliminary}, employing a deterministic DDIM \cite{ddim} sampler to recover 3D poses from pure Gaussian noise. Specifically, we iteratively refine $H$ initial hypotheses over $K$ steps and subsequently aggregate these diverse predictions to ensure robust reconstruction.

\subsection{Temporal Correlation-Enhanced Pruning (TCEP)}
\label{tcep}

Learning fine-grained temporal correlations is crucial for accurate 3D human pose estimation, particularly in diffusion-based frameworks where iterative denoising must preserve subtle motion cues. However, constructing dense temporal connections for each pose token risks computational inefficiency and semantic drift. To address this within the \textit{Frame-level Pruning} phase, we design the TCEP module to establish the structural foundation of our hierarchy. By explicitly modeling sparse temporal dependencies, TCEP constructs a preliminary sparse topology by dynamically selecting semantically correlated frames for each joint while filtering out irrelevant connections, thereby ensuring that subsequent feature learning is strictly focused on motion-critical regions.

Given the input $\boldsymbol{Y}_{t}$, TCEP first overlays the learnable global temporal topology matrix $\mathbf{\hat{A}}_{F} \in\mathbb{R}^{F \times F}$ onto the original adjacency matrix $\mathbf{A}_{F} \in\mathbb{R}^{F \times F}$, as outlined by \cite{agformer, ktpformer}. This combination can be expressed as:

\begin{equation}
\mathbf{A}_T=\frac{(\mathbf{A}_F+\mathbf{\hat{A}}_F)+(\mathbf{A}_F+\mathbf{\hat{A}}_F)'}{2},
\end{equation}
where $'$ denotes the matrix transpose to ensure symmetry. As illustrated in Fig. \ref{fig_architecture} top left, we apply a Correlation-Enhanced Node Selecting algorithm to dynamically select temporal correlation nodes and refine their connections. For each joint $j\in\{1,\ldots,J\}$, $\boldsymbol{Y}_t^{(j)}\in\mathbb{R}^{F\times D}$ denotes its temporal token sequence. We compute a scaled similarity matrix:

\begin{equation}\label{eq6}
\mathbf{S}^{(j)}=\frac{\boldsymbol{Y}_t^{(j)}\left(\boldsymbol{Y}_t^{(j)}\right)^\top}{\sqrt{D}}\in\mathbb{R}^{F\times F},
\end{equation}
where each entry $\mathbf{S}^{(j)}_{pq}$ quantifies the correlation between frame $p$ and $q$. 

To sparsify the topology, we suppress the diagonal (\ie, self-attention) and retain only the top-$\eta$ highest-scoring off-diagonal neighbors for each row $p$, forming the directed neighborhood set $\mathrm{Top}_p(\mathbf{S}^{(j)}, \eta) \subset \{1, \ldots, F\}$. These selections are then used to construct a binary mask $\mathbf{M}^{(j)} \in \{0, 1\}^{F \times F}$ by restoring self-loops and applying symmetric completion:

\begin{equation} \label{eq7}
\mathbf{M}^{(j)}_{pq}=\begin{cases}1,&p=q~\mathrm{or}~q/p\in\mathrm{Top}_{p/q}(\mathbf{S}^{(j)},\eta),\\0,&\text{otherwise.}\end{cases}
\end{equation}

The detailed procedure for generating this joint-wise symmetric mask is summarized in Algorithm \ref{alg:tcep_mask} Phase 1. To enforce sparse support, we update the similarity matrix by setting non-selected entries to a large negative constant, forming the masked version $\check{\mathbf{S}}^{(j)}$ that satisfies:

\begin{equation}
\mathbf{\check{S}}^{(j)}_{pq}=\begin{cases}\mathbf{S}^{(j)}_{pq},& \text{if}~ \mathbf{M}^{(j)}_{pq}=1,\\-\infty,&\text{otherwise,}\end{cases}
\end{equation}
this masked matrix ensures that only top-$\eta$ correlations are retained during the subsequent softmax normalization.

The masked similarity $\check{\mathbf{S}}^{(j)}$ is normalized using the softmax function, and fused with the global adjacency $\mathbf{A}_T$ via Hadamard product to yield a joint-specific attention matrix:

\begin{equation}
\mathbf{\check{A}}^{(j)}_T=\mathbf{A}_{T} \odot \sigma_1(\mathbf{\check{S}}^{(j)}).
\end{equation}

Finally, the pruned temporal graph is used to refine joint-wise tokens through linear projection and nonlinear activation with residual connection:

\begin{equation}
\boldsymbol{Y}^{\prime(j)}_t=\boldsymbol{Y}^{(j)}_{t} + \sigma_2(\mathbf{\check{A}}^{(j)}_T \boldsymbol{Y}^{(j)}_{t}\odot\mathbf{W}) \in \mathbb{R}^{F\times D},
\end{equation}
where $\mathbf{W} \in \mathbb{R}^{D \times D}$ is a shared linear layer, and $\sigma_2(\cdot)$ is the GELU activation function.

We then stack $\{\boldsymbol{Y}_t^{\prime(j)}\}_{j=1}^J$ and $\{\mathbf{M}^{(j)}\}_{j=1}^J$ to reconstruct $\boldsymbol{Y}^{\prime}_t \in \mathbb{R}^{J \times F \times D}$ and the final temporal mask $\mathbf{M} \in \{0, 1\}^{J \times F \times F}$ for downstream modules.

\begin{table*}[ht]
    \centering
    \caption{\textbf{Quantitative Comparison with SOTA Methods on the Human3.6M Dataset Under MPJPE for Various Actions.} $Dir., Disc., \cdots,$ and $WalkT.$ Correspond to 15 Action Classes. $Avg$ Indicates the Average MPJPE Among 15 Action Classes. The Best and Second-Best Results Are Highlighted in \textbf{Bold} and \underline{Underline} Formats.}
    \renewcommand{\arraystretch}{1.2}
    \resizebox{1.0\textwidth}{!}
    {
    \begin{tabular}{l c| *{15}{c}|c}
        \toprule
        \multirow[c]{2}{*}{Method} & \multirow[c]{2}{*}{Type} & \multicolumn{16}{c}{MPJPE $\downarrow$} \\
        \cmidrule(lr){3-18}
        & & Dir. & Disc. & Eat & Greet & Phone & Photo & Pose & Pur. & Sit & SitD. & Smoke & Wait & WalkD. & Walk & WalkT. & Avg. \\
        \midrule
        TCN~\cite{tcn} & CNN & 45.2 & 46.7 & 43.3 & 45.6 & 48.1 & 55.1 & 44.6 & 44.3 & 57.3 & 65.8 & 47.1 & 44.0 & 49.0 & 32.8 & 33.9 & 46.8 \\
        DUE~\cite{due} & GCN & 37.9 & 41.9 & 36.8 & 39.5 & 40.8 & 49.2 & 40.1 & 40.7 & 47.9 & 53.3 & 40.2 & 41.1 & 40.3 & 30.8 & 28.6 & 40.6\\
        GLA-GCN~\cite{gla-gcn} & GCN & 41.3 & 44.3 & 40.8 & 41.8 & 45.9 & 54.1 & 42.1 & 41.5 & 57.8 & 62.9 & 45.0 & 42.8 & 45.9 & 29.4 & 29.9 & 44.4 \\
        FTCM~\cite{ftcm} & MLP & 42.2 & 44.4 & 42.4 & 42.4 & 47.7 & 55.8 & 42.7 & 41.9 & 58.7 & 64.5 & 46.1 & 44.2 & 45.2 & 30.6 & 31.1 & 45.3 \\ 
        \midrule
        PoseFormer~\cite{poseformer} & Transformer & 41.5 & 44.8 & 39.8 & 42.5 & 46.5 & 51.6 & 42.1 & 42.0 & 53.3 & 60.7 & 45.5 & 43.3 & 46.1 & 31.8 & 32.2 & 44.3 \\ 
        GraFormer~\cite{graformer} & Transformer & 45.2 & 50.8 & 48.0 & 50.0 & 54.9 & 65.0 & 48.2 & 47.1 & 60.2 & 70.0 & 51.6 & 48.7 & 54.1 & 39.7 & 43.1 & 51.8 \\
        P-STMO~\cite{p-stmo} & Transformer  & 38.9 & 42.7 & 40.4 & 41.1 & 45.6 & 49.7 & 40.9 & 39.9 & 55.5 & 59.4 & 44.9 & 42.2 & 42.7 & 29.4 & 29.4 & 42.8 \\
        MixSTE~\cite{mixste} & Transformer & 36.7 & 39.0 & 36.5 & 39.4 & 40.2 & 44.9 & 39.8 & 36.9 & 47.9 & 54.8 & 39.6 & 37.8 & 39.3 & 29.7 & 30.6 & 39.8 \\
        MHFormer~\cite{mhformer} & Transformer & 39.2 & 43.1 & 40.1 & 40.9 & 44.9 & 51.2 & 40.6 & 41.3 & 53.5 & 60.3 & 43.7 & 41.1 & 43.8 & 29.8 & 30.6 & 43.0 \\
        STCFormer~\cite{stcformer} & Transformer & 38.4 & 41.2 & 36.8 & 38.0 & 42.7 & 50.5 & 38.7 & 38.2 & 52.5 & 56.8 & 41.8 & 38.4 & 40.2 & 26.2 & 27.7 & 40.5 \\
        MotionBERT~\cite{motionbert}$^{\ddagger}$ & Transformer & 36.1 & 37.5 & 35.8 & 32.1 & 40.3 & 46.3 & 36.1 & 35.3 & 46.9 & 53.9 & 39.5 & 36.3 & 35.8 & 25.1 & 25.3 & 37.5 \\
        TC-MixSTE~\cite{tc-mixste} & Transformer & 36.9 & 40.9 & 36.3 & 39.0 & 41.6 & 48.7 & 38.4 & 39.3 & 50.3 & 54.9 & 40.6 & 38.0 & 40.6 & 26.5 & 26.0 & 39.9 \\
        DualFormer~\cite{dualformer} &Transformer & 38.9 & 43.1 & 39.2 & 41.4 & 45.1 & 50.7 & 41.5 & 41.2 & 51.7 & 60.5 & 43.4 & 41.4 & 43.0 & 29.9 & 31.0 & 42.8 \\
        \midrule
        Diffpose~\cite{diffpose} & Diffusion & 33.2 & 36.6 & 33.0 & 35.6 & 37.6 & 45.1 & 35.7 & 35.5 & 46.4 & 49.9 & 37.3 & 35.6 & 36.5 & 24.4 & 24.1 & 36.9 \\
        D3DP~\cite{d3dp} & Diffusion & 33.0 & 34.8 & 31.7 & 33.1 & 37.5 & 43.7 & 34.8 & 33.6 & 45.7 & 47.8 & 37.0 & 35.0 & 35.0 & 24.3 & 24.1 & 35.4 \\
        FinePOSE~\cite{finepose} & Diffusion & 31.4 & \underline{31.5} & \underline{28.8} & \underline{29.7} & \underline{34.3} & \underline{36.5} & \underline{29.2} & \underline{30.0} & \underline{42.0} & \underline{42.5} & \underline{33.3} & \underline{31.9} & \underline{31.4} & 22.6 & \underline{22.7} & \underline{31.9} \\
        KTPFormer~\cite{ktpformer} & Diffusion & \underline{30.1} & 32.1 & 29.1 & 30.6 & 35.4 & 39.3 & 32.8 & 30.9 & 43.1 & 45.5 & 34.7 & 33.2 & 32.7 & \underline{22.1} & 23.0 & 33.0 \\
        \rowcolor{black!10} HTP (Ours) & Diffusion & \textbf{28.5} & \textbf{30.0} & \textbf{26.4} & \textbf{27.2} & \textbf{31.5} & \textbf{36.0} & \textbf{28.8} & \textbf{27.7}  & \textbf{39.5} & \textbf{39.1} & \textbf{30.7}  & \textbf{29.1} & \textbf{30.7} & \textbf{21.7} & \textbf{22.3} & \textbf{29.9} \\  
        \bottomrule
    \end{tabular}    
    }
    \label{tab:h36m_res2}
\end{table*}

\begin{table*}[ht]
    \centering
    \caption{\textbf{Quantitative Comparison with SOTA Methods on the Human3.6M Dataset Under P-MPJPE for Various Actions.} The Best and Second-Best Results Are Highlighted in \textbf{Bold} and \underline{Underline} Formats.}
    \renewcommand{\arraystretch}{1.2}
    \resizebox{1.0\textwidth}{!}
    {
    \begin{tabular}{l c| *{15}{c}|c}
        \toprule
        \multirow[c]{2}{*}{Method} & \multirow[c]{2}{*}{Type} & \multicolumn{16}{c}{P-MPJPE $\downarrow$} \\
        \cmidrule(lr){3-18}
        & & Dir. & Disc. & Eat & Greet & Phone & Photo & Pose & Pur. & Sit & SitD. & Smoke & Wait & WalkD. & Walk & WalkT. & Avg. \\
        \midrule
        TCN~\cite{tcn} & CNN & 34.1 & 36.1 & 34.4 & 37.2 & 36.4 & 42.2 & 34.4 & 33.6 & 45.0 & 52.5 & 37.4 & 33.8 & 37.8 & 26.6 & 27.3 & 36.5 \\
        DUE~\cite{due} & GCN & 30.3 & 34.6 & 29.6 & 31.7 & 31.6 & 38.9 & 31.8 & 31.9 & 39.2 & 42.8 & 32.1 & 32.6 & 31.4 & 25.1 & 23.8 & 32.5\\
        GLA-GCN~\cite{gla-gcn} & GCN & 32.4 & 35.3 & 32.6 & 34.2 & 35.0 & 42.1 & 32.1 & 31.9 & 45.5 & 49.5 & 36.1 & 32.4 & 35.6 & 23.5 & 24.7 & 34.8 \\
        FTCM~\cite{ftcm} & MLP & 31.9 & 35.1 & 34.0 & 34.2 & 36.0 & 42.1 & 32.3 & 31.2 & 46.6 & 51.9 & 36.5 & 33.8 & 34.4 & 23.8 & 24.9 & 35.3 \\
        \midrule
        PoseFormer~\cite{poseformer} & Tansformer & 34.1 & 36.1 & 34.4 & 37.2 & 36.4 & 42.2 & 34.4 & 33.6 & 45.0 & 52.5 & 37.4 & 33.8 & 37.8 & 25.6 & 27.3 & 36.5 \\ 
        P-STMO~\cite{p-stmo} & Tansformer  & 31.3 & 35.2 & 32.9 & 33.9 & 35.4 & 39.3 & 32.5 & 31.5 & 44.6 & 48.2 & 36.3 & 32.9 & 34.4 & 23.8 & 23.9 & 34.4 \\
        MixSTE~\cite{mixste} & Tansformer & 30.8 & 33.1 & 30.3 & 31.8 & 33.1 & 39.1 & 31.1 & 30.5 & 42.5 & 44.5 & 34.0 & 30.8 & 32.7 & 22.1 & 22.9 & 32.6 \\
        MHFormer~\cite{mhformer} & Tansformer & 31.5 & 34.9 & 32.8 & 33.6 & 35.3 & 39.6 & 32.0 & 32.2 & 43.5 & 48.7 & 36.4 & 32.6 & 34.3 & 23.9 & 25.1 & 34.4 \\
        STCFormer~\cite{stcformer} & Tansformer & 29.3 & 33.0 & 30.7 & 30.6 & 32.7 & 38.2 & 29.7 & 28.8 & 42.2 & 45.0 & 33.3 & 29.4 & 31.5 & 20.9 & 22.3 & 31.8 \\
        TC-Mixste~\cite{tc-mixste} & Transformer & 29.5 & 32.8 & 28.9 & 31.6 & 32.8 & 37.9 & 29.8 & 29.1 & 41.8 & 44.3 & 33.5 & 30.6 & 32.2 & 21.2 & 22.2 & 31.9 \\
        DualFormer~\cite{dualformer} &Transformer & 31.4 & 34.9 & 32.5 & 34.3 & 35.1 & 39.4 & 33.0 & 32.0 & 42.9 & 48.9 & 36.2 & 32.9 & 33.5 & 23.7 & 25.1 & 34.4 \\
        \midrule
         D3DP~\cite{d3dp} & Diffusion & 27.5 & 29.4 & 26.6 & 27.7 & 29.2 & 34.3 & 27.5 & 26.2 & 37.3 & 39.0 & 30.3 & 27.7 & 28.2 & 19.6 & 20.3 & 28.7 \\
        KTPFormer~\cite{ktpformer} & Diffusion & \underline{24.1} & \underline{26.7} & \underline{24.2} & \underline{24.9} & \underline{27.3} & \underline{30.6} & \underline{25.2} & \underline{23.4} & \underline{34.1} & \underline{35.9} & \underline{28.1} & \underline{25.3} & \underline{25.9} & \underline{17.8} & \underline{18.8} & \underline{26.2} \\
        \rowcolor{black!10} HTP (Ours) & Diffusion & \textbf{21.8} & \textbf{23.4} & \textbf{21.2} & \textbf{21.9} & \textbf{23.8} & \textbf{27.9} & \textbf{21.9} & \textbf{20.8}  & \textbf{31.2} & \textbf{31.2} & \textbf{24.6}  & \textbf{21.8} & \textbf{23.5} & \textbf{17.3} & \textbf{17.4} & \textbf{23.3} \\
        \bottomrule
    \end{tabular}    
    }
    \label{tab:h36m_res3}
\end{table*}

\subsection{Sparse-Focused Temporal MHSA (SFT MHSA)}
\label{sft-mhsa}

Serving as an intermediate refinement within the \textit{Frame-level Pruning} phase, we introduce the SFT MHSA module. Guided by the sparse temporal mask $\mathbf{M}$ generated by TCEP, it restricts attention computations to semantically correlated frames. Importantly, the module functions as a semantic bridge, enhancing the distinctiveness of the selected frames and ensuring that the tokens are semantically robust before undergoing physical compression in the next phase.

The architecture of the SFT MHSA module is illustrated in Fig.~\ref{fig_architecture} (b). Given the temporally correlated pose token sequence $\boldsymbol{Y}^{\prime}_t \in \mathbb{R}^{J \times F \times D}$, we first apply a LayerNorm and project the features into multi-head attention components: query ($\boldsymbol{Q}$), key ($\boldsymbol{K}$), and value ($\boldsymbol{V}$). These are obtained via linear projections using learnable weight matrices $\boldsymbol{W}_Q$, $\boldsymbol{W}_K$, and $\boldsymbol{W}_V \in \mathbb{R}^{D \times D}$, respectively. The multi-head self-attention computation is then performed in parallel across $h$ attention heads, each with dimension $d_k = D/h$. As shown in Algorithm \ref{alg:tcep_mask} Phase 2, to enforce sparsity, we define an additive mask $\mathbf{M}' \in \mathbb{R}^{J \times F \times F}$ constructed by applying the following rule to each joint-specific binary mask $\mathbf{M}^{(j)} \in \{0,1\}^{F \times F}$:

\begin{equation} \label{eq11}
\mathbf{M}^{\prime(j)}_{pq} =
\begin{cases}
0 ,& \text{if } \mathbf{M}^{(j)}_{pq} = 1, \\
-\infty ,& \text{if } \mathbf{M}^{(j)}_{pq} = 0,
\end{cases}
\end{equation}
where $p,q \in \{1, \ldots, F\}$. These joint-level masks are stacked to obtain $\mathbf{M}'$, ensuring that only relevant temporal connections are preserved during attention computation, while others are suppressed by assigning near-zero weights via softmax. The attention for each head is then computed as:

\begin{equation}
\text{head}_i = \text{softmax}\left( \frac{\boldsymbol{Q}_i \boldsymbol{K}_i^T}{\sqrt{d_k}} + \mathbf{M}' \right) \boldsymbol{V}_i,
\end{equation}
where $\boldsymbol{Q}_i, \boldsymbol{K}_i, \boldsymbol{V}_i \in \mathbb{R}^{J \times F \times d_k}$ are the $i$-th head-specific projections. The outputs of all heads are concatenated and projected via $\boldsymbol{W}_O \in \mathbb{R}^{D \times D}$, followed by a residual connection:

\begin{equation}
\boldsymbol{\tilde{Y}}_t = \text{Concat}(\text{head}_1, \dots, \text{head}_h) \boldsymbol{W}_O + \boldsymbol{Y}^{\prime}_t.
\end{equation}

The result is further processed through a LayerNorm-MLP block with GELU activation, and finalized with a second residual connection:

\begin{equation}
\boldsymbol{\tilde{Y}}^{\prime}_t = \text{MLP}(\text{LN}(\boldsymbol{\tilde{Y}}_t)) + \boldsymbol{\tilde{Y}}_t.
\end{equation}

\begin{figure}[t]
\centering
\includegraphics[width=1\columnwidth]{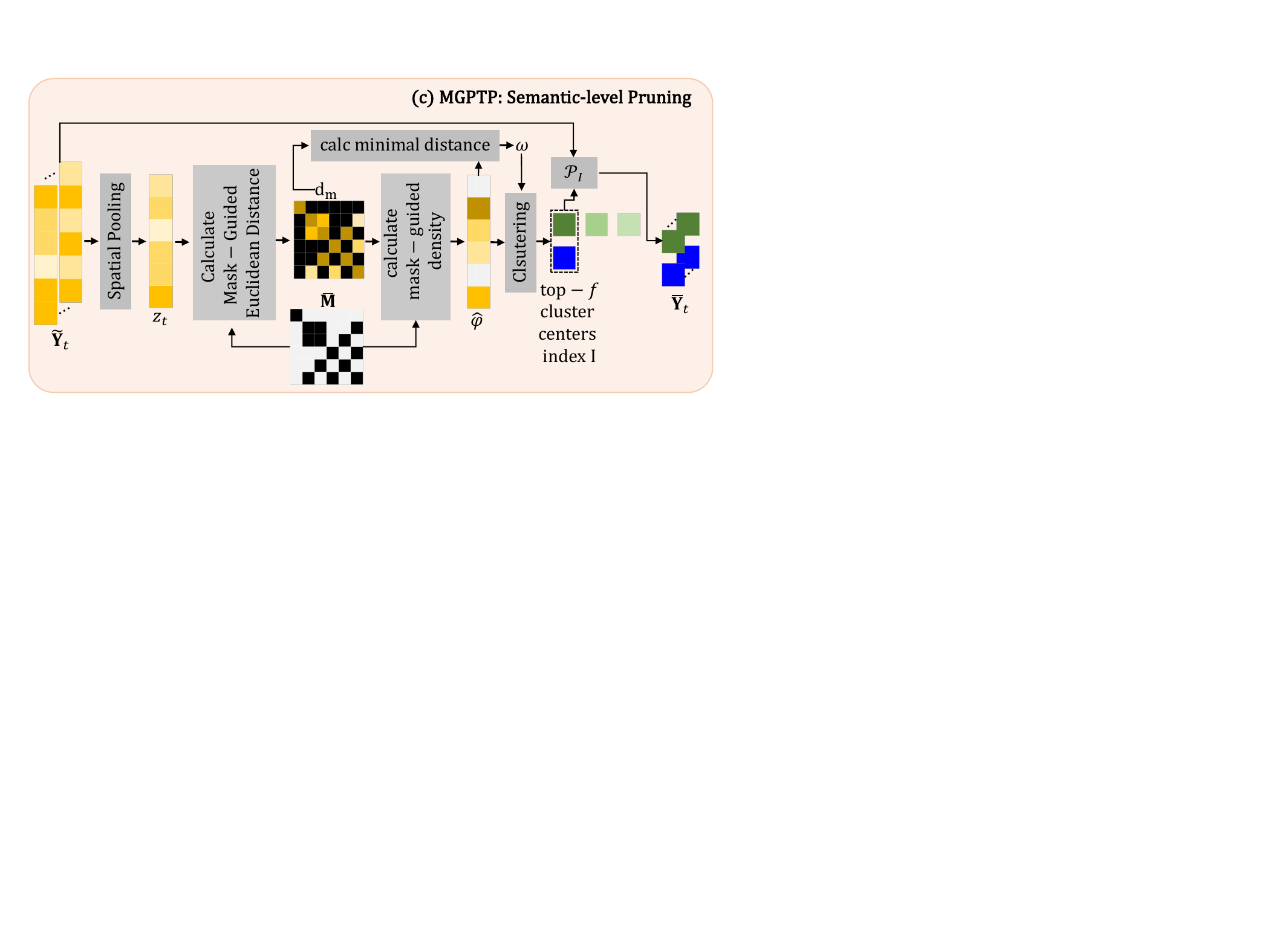}
\caption{The architecture of the MGPTP.}
\label{fig_mgptp}
\end{figure}

\subsection{Mask-Guided Pose Token Pruner (MGPTP)}
\label{mgptp}

Advancing to the \textit{Semantic-level Pruning} phase, we propose the MGPTP module to transform the refined frame-level features into a compact semantic representation. Operating on the temporally refined tokens from SFT MHSA, MGPTP dynamically selects a compact subset of semantically informative frames by leveraging the learned attention masks. This mechanism aggregates temporal context into high-level semantic descriptors, maximizing computational efficiency while preserving essential motion cues.

As shown in Fig.~\ref{fig_mgptp}, given the pose tokens after SFT MHSA, $\boldsymbol{\Tilde{Y}}^{\prime}_{t} \in \mathbb{R}^{J \times F \times D}$, we apply average pooling along the joint dimension to obtain frame-wise tokens $\mathbf{z}_t \in \mathbb{R}^{F \times D}$. Simultaneously, the joint-wise temporal mask $\mathbf{M} \in \{0,1\}^{J \times F \times F}$ is aggregated across joints via the same pooling operation, producing a smoothed frame-wise attention mask $\overline{\mathbf{M}} \in \mathbb{R}^{F \times F}$. This unified step ensures both the token representation and its corresponding temporal guidance share consistent semantic granularity across frames.

Then, a novel mask-guided density peaks clustering based on the k-nearest neighbors algorithm is employed to cluster pose tokens with high motion relevance, guided by $\overline{\mathbf{M}}$. Specifically, for any pair of frame tokens $\mathbf{z}_{p},\mathbf{z}_{q}  \in \mathbf{z}_{t}$, the mask-guided euclidean distance is defined as:

\begin{equation}\mathrm{d}_{\mathrm{m}}(\mathbf{z}_p,\mathbf{z}_q)=\begin{cases}\frac{\|\mathbf{z}_p-\mathbf{z}_q\|_2}{\sqrt{D}},
&\overline{\mathbf{M}}_{pq}=1,\\\Lambda,&\overline{\mathbf{M}}_{pq}=0,\end{cases}
\end{equation}
where $\Lambda=\max_{p,q}\|\mathbf{z}_p-\mathbf{z}_q\|_2/\sqrt{D}+\varepsilon$. $\varepsilon>0$ is a small constant ensuring that pairs masked out by $\overline{\mathbf{M}}$ are strictly farther than any valid pair, effectively excluding them during neighborhood search.

Let $k$ be a fixed neighborhood size, and $\mathrm{NN}_k(\mathbf{z}_{p})$ be the $k$-th nearest point to $\mathbf{z}_{p}$ according to $\mathrm{d_{m}}$. Thus, the mask-guided k-nearest neighbors $\mathrm{KNN}_{\mathrm{m}}(\cdot)$ of $\mathbf{z}_{p}$ is defined as:

\begin{equation}
\mathrm{KNN}_{m}(\mathbf{z}_{p})=\left\{\mathbf{z}_{q}\in \mathbf{z}_{t}|\mathbf{d}_{m}(\mathbf{z}_{p},\mathbf{z}_{q})\leq\mathbf{d}_{m}(\mathbf{z}_{p},\mathrm{NN}_k(\mathbf{z}_{p}))\right\}.
\end{equation}

The local density $\varphi_p$ of frame $p$ is then computed via a Guassian kernel over $\mathrm{KNN}_{\mathrm{m}}(\mathbf{z}_{p})$:

\begin{equation}
\varphi_p=\exp\left(-\frac{1}{k}\sum_{\mathbf{z}_{q}\in \mathrm{KNN}_{\mathrm{m}}(\mathbf{z}_{p})} \mathrm{d_{m}}(\mathbf{z}_{p},\mathbf{z}_{q})^2\right).
\end{equation}

To emphasize frames that maintain richer temporal connectivity, we aggregate mask support per frame $s_p=\sum_{q=1}^F\overline{\mathbf{M}}_{pq}$, apply a stability-aware transformation:

\begin{equation}
\tilde{s}_p=\begin{cases}s_p,&\text{if } s_p>0,\\-\infty,&\text{otherwise.}\end{cases}
\end{equation}

The mask-guided response density $\hat{\varphi}_{p}$ is then computed by:

\begin{equation}
\hat{\varphi}_{p} = \varphi_p \cdot\sigma_1\left(\tilde{s}_p\right),
\end{equation}
where $\sigma_1(\cdot)$ denotes the softmax function applied across all $\tilde{s}_p$. Next, the minimal distance $\omega_p$ to higher-density neighbors is computed for each frame:

\begin{equation}
\omega_p =
\begin{cases}
    \min_{q : \hat{\varphi}_{q} > \hat{\varphi}_{p}} \mathrm{d_{m}}(\mathbf{z}_{p}, \mathbf{z}_{q}), & \text{if } \exists \hat{\varphi}_{q} > \hat{\varphi}_{p}, \\
    \max_{q} \mathrm{d_{m}}(\mathbf{z}_{p}, \mathbf{z}_{q}), & \text{otherwise.}
\end{cases}
\end{equation}

Following the clustering principle in \cite{dpc_knn}, the saliency score of each token $\mathbf{z}_{p}$ is defined as $\omega_p \times \hat{\varphi}_p$. We then select the top-$f$ cluster centers and obtain an ordered index set $I=\{i_1<i_2<\cdots<i_f\}$ on the temporal axis. Finally, we define an order-preserving temporal selection operator $\mathcal{P}_I$ applied to the original joint-wise token sequence, which extracts the temporal slices at positions $I$ uniformly across all joints. The resulting sequence is

\begin{equation}
\bar{\boldsymbol{Y}}_t=\mathcal{P}_I(\boldsymbol{\Tilde{Y}}^{\prime}_{t})\in\mathbb{R}^{J\times f\times D},
\end{equation}
which retains the full spatial and feature dimensions while compressing the temporal axis. This pruned sequence maintains temporal coherence and semantic consistency across all joints, enabling more efficient downstream reasoning.

\section{Experiments}
\label{sec:experiments}

\subsection{Datasets and Metrics}

\textbf{\textit{1) Human3.6M}\cite{human3.6m}:} Human3.6M is the largest and most widely used indoor benchmark dataset for HPE tasks, comprising 3.6 million RGB images covering 15 activities performed by 11 actors. Videos are recorded at 50Hz using four synchronized and calibrated cameras. Following \cite{mhformer, mixste, d3dp} , our model is trained on 5 subjects (S1, S5, S6, S7, S8) and evaluated on 2 subjects (S9, S11). For evaluation metrics, we report the mean per joint position error (MPJPE) and Procrustes MPJPE (P-MPJPE).

\textbf{\textit{2) MPI-INF-3DHP} \cite{3dhp}:} MPI-INF-3DHP is a recently popular dataset consisting of indoor and outdoor scenes. The training set contains 8 activities performed by 8 actors, while the test set covers 7 activities. Following the protocol in \cite{p-stmo}, we use the area under the curve (AUC), percentage of correct keypoints (PCK), and MPJPE as evaluation metrics. 

\subsection{Implementation Details}
\textbf{\textit{1) Training Details}:} We use CPN \cite{cpn} as the 2D keypoint detector to generate the 2D inputs. The numbers of SFT MHSA $n_{1}$ are set to 3. The batch size is set to 4, with each sample containing a pose sequence of 243 frames. For temporal pruning, the pruning length $f$ is selected based on dataset-specific temporal characteristics. For Human3.6M, which contains long and high-frame-rate sequences with substantial temporal redundancy, a pruning length of $f=54$ provides an effective balance between removing redundant frames and retaining sufficient motion granularity. For MPI-INF-3DHP, where sequences are shorter and motion cues are more concentrated, a conservative pruning ratio of 3:1 ($f=27$) is adopted to preserve essential fine-grained motion information. We adopt the AdamW \cite{adamw} optimizer with the momentum parameters of $\beta_{1}, \beta_{2} = 0.9, 0.999$, and a weight decay of 0.1. We train our model for 150 epochs and the initial learning rate is $\text{6e}^{-5}$ with a shrink factor of 0.993 after each epoch. For fair comparisons, we set the number of hypotheses $H=1$ and iterations $K=1$ during training, and $H=20$ and $K=10$ during inference, as in D3DP \cite{d3dp}.

\textbf{\textit{2) Implementation of Plug-and-Play Integration}:} To verify generality, we integrated HTP into Transformer-based frameworks. All variants were trained from scratch to ensure fair comparison. For MixSTE~\cite{mixste}, the integration follows the identical architectural modifications as our main method. For the dual-stream MotionBERT~\cite{motionbert}, we inserted TCEP and an initial SFT MHSA layer prior to the backbone. Inside the first $n_1=2$ DSTformer blocks, we replaced the original Temporal MHSA components with our SFT MHSA. The MGPTP module was inserted after the second block, allowing the remaining layers to process the efficient, pruned sequence.

\textbf{\textit{3) Implementation of Efficiency Evaluation}:} We report the total Multiply-Accumulate operations (MACs), representing the cumulative computational cost aggregated across all $K$ denoising iterations and $H$ hypotheses. These values are computed via the THOP library, following the standard convention where $1 \text{ MACs} \approx 2 \text{ FLOPs}$. Inference speed (FPS) is measured on two NVIDIA GeForce RTX 4090 GPUs with batch size of 8 under FP32 precision. The FPS is calculated by dividing the total number of processed frames by the total wall-clock time of inference, excluding data loading overhead.

\subsection{Quantitative Results}
\textbf{\textit{1) Human3.6M}:} Tab. \ref{tab:h36m_res1} presents comparisons between our HTP and recent state-of-the-art (SOTA) 3D HPE methods on the Human3.6M dataset. Our approach achieves SOTA performance, with MPJPE of $29.9\text{mm}$ and P-MPJPE of $23.3\text{mm}$ using 2D poses detected by CPN \cite{cpn} as inputs, and MPJPE of $16.7\text{mm}$ when using ground-truth 2D poses as inputs. With an $81\%$ reduction in parameter count and a $40.0\%$ decrease in computational cost, HTP surpasses the previous SOTA method FinePose \cite{finepose} by $2.0\text{mm}$ in MPJPE and $1.7\text{mm}$ in P-MPJPE. 

Additionally, to demonstrate the plug-and-play capability of our design, we integrate the proposed Hierarchical Temporal Pruning strategy into two Transformer-based frameworks: MixSTE \cite{mixste} and MotionBERT \cite{motionbert}. On MixSTE, HTP reduces MACs by $37\%$ while improving MPJPE by $1.0\text{mm}$ and P-MPJPE by $0.7\text{mm}$. On MotionBERT, we observe similar gains, with a $42\%$ reduction in MACs and improved pose accuracy across both metrics. We further compare HTP with emerging Mamba-based methods, PoseMamba-X~\cite{posemamba} and SAMA-L~\cite{sama}. While these state-space models offer competitive efficiency, HTP demonstrates superior reconstruction fidelity, outperforming SAMA-L by 7.0 mm in MPJPE.

Further, Tab.\ref{tab:h36m_res2} and Tab.\ref{tab:h36m_res3} report per-action results on the Human3.6M dataset under MPJPE and P-MPJPE metrics, using detected 2D keypoints \cite{cpn} as input. Our HTP framework consistently achieves the lowest error across all 15 action categories, outperforming all prior state-of-the-art methods. Notably, significant improvements are observed in challenging categories such as \textit{``SitD"}, \textit{``Walk"}, and \textit{``WalkT"}, highlighting its ability to handle diverse motion dynamics with precision. 


\begin{table}[t]
    \centering
    \caption{\textbf{Quantitative Comparison with SOTA Methods on the MPI-INF-3DHP Dataset.} The Best and Second-Best Results Are Highlighted in Bold and \underline{Underline} Formats.}
    \renewcommand{\arraystretch}{1.2}
    \resizebox{1.0\linewidth}{!}
    {
    \begin{tabular}{l c|c ccc}
        \toprule
        \multirow[c]{2}{*}{Method} & \multirow[c]{2}{*}{Type} & \multirow[c]{2}{*}{$F$} & \multicolumn{3}{c}{MPI-INF-3DHP} \\
        \cmidrule(lr){4-6}
        & & & PCK $\uparrow$ & AUC $\uparrow$ & MPJPE $\downarrow$\\
        \midrule
        TCN~\cite{tcn} & CNN & 81 & 86.0 & 51.9 & 84.0 \\
        GLA-GCN~\cite{gla-gcn} & GCN  & 81 & 98.5 & 79.1 & 27.8  \\
        FTCM~\cite{ftcm} & MLP & 81 & 98.0 & 79.8 & 31.2 \\
        \midrule
        PoseFormer~\cite{poseformer} & Transformer & 9 & 88.6 & 56.4 & 77.1 \\ 
        P-STMO~\cite{p-stmo} & Transformer  & 81  & 97.9 & 75.8 & 32.2  \\
        MixSTE~\cite{mixste} & Transformer  & 27  & 94.4 & 66.5 & 54.9  \\
        PoseFormerV2~\cite{poseformerv2} & Transformer & 81 & 97.9 & 78.8 & 27.8  \\
        MHFormer~\cite{mhformer} & Transformer & 9 & 93.8 & 63.3 & 58.0  \\
        TC-MixSTE~\cite{tc-mixste} & Transformer & 81 & 98.7 & 79.5 & 27.6 \\
        DualFormer~\cite{dualformer} & Transformer & 9 & 97.8 & 73.4 & 40.1 \\
        \midrule
        Diffpose~\cite{diffpose} & Diffusion & 81 & 98.0 & 75.9 & 29.1  \\
        D3DP~\cite{d3dp} & Diffusion & 81 & 98.0 & 79.1 & 28.1  \\
        KTPFormer~\cite{ktpformer} & Diffusion & 81 & \underline{99.0} & 79.3 & 29.1  \\ 
        FinePOSE~\cite{finepose} & Diffusion & 81 & 98.9 & \underline{80.0} & \textbf{26.2}  \\
        \rowcolor{black!10} HTP (Ours) & Diffusion & 81 & \textbf{99.5} & \textbf{80.5} & \underline{26.4} \\
        \bottomrule
    \end{tabular}  
    }
    \label{tab:3dhp_res}
\end{table}

\textbf{\textit{2) MPI-INF-3DHP}:} Tab. \ref{tab:3dhp_res} presents comparisons between our method and recent SOTA 3D HPE approaches on the MPI-INF-3DHP dataset. Following \cite{d3dp, finepose}, our model is trained using ground-truth 2D poses as inputs. Compared to recent SOTA works \cite{finepose, ktpformer}, HTP maintains comparable MPJPE while improving PCK by $0.5\%$ and AUC by $0.5\%$. While the performance gain is less pronounced than in Human3.6M due to the reduced temporal redundancy available for pruning in shorter sequences, these results confirm that HTP remains robust and effective even under constrained temporal resolutions.

\begin{table}[t]
    \centering
    \caption{\textbf{MACs and Speed Comparison with Diffusion-Based 3D HPE Methods.} All Models Are Evaluated Under the Same Setting. Best Results Are \textbf{Bolded}.}
    \renewcommand{\arraystretch}{1.2}
    \resizebox{1.0\linewidth}{!}
    {
    \begin{tabular}{l|cc|c|cc}
        \toprule
        \multirow{2}{*}{\textbf{Method}} & 
        \multirow{2}{*}{MPJPE$\downarrow$} & 
        \multirow{2}{*}{Params (M)} & 
        \textbf{Train} & 
        \multicolumn{2}{c}{\textbf{Inference}} \\
        \cmidrule(lr){4-6}
        & & & MACs/frame & MACs/frame & FPS$\uparrow$ \\
        \midrule
        \rowcolor{gray!10} \multicolumn{6}{l}{\textbf{\textit{Setting:} $K = 1$, $H = 20$}} \\
        D3DP~\cite{d3dp}         & 38.8 & 34.8  & 0.58 G & 22.9 G & 772.7 \\
        FinePose~\cite{finepose} & 40.0 & 200.6 & 0.60 G & 22.9 G & 723.7 \\
        KTPFormer~\cite{ktpformer} & 39.5 & 36.3  & 0.58 G & 23.6 G & 705.8 \\
        \rowcolor{black!10} 
        HTP (Ours) & \textbf{32.9} & 37.5 & \textbf{0.36 G} & \textbf{10.0 G} & \textbf{2277.5} \\
        \midrule
        \rowcolor{gray!10} \multicolumn{6}{l}{\textbf{\textit{Setting:} $K = 10$, $H = 20$}} \\
        D3DP~\cite{d3dp}         & 35.4 & 34.8  & 0.58 G & 228.8 G & 79.6 \\
        FinePose~\cite{finepose} & 31.9 & 200.6 & 0.60 G & 236.2 G & 73.8 \\
        KTPFormer~\cite{ktpformer} & 33.0 & 36.3  & 0.58 G & 228.8 G & 73.5 \\
        \rowcolor{black!10} 
        HTP (Ours) & \textbf{29.9} & 37.5 & \textbf{0.36 G} & \textbf{99.8 G} & \textbf{137.0} \\
        \bottomrule
    \end{tabular}
    }
    \label{tab:efficiency_comparison}
\end{table}

\textbf{\textit{3) MACs and Speed}:}
Tab.~\ref{tab:efficiency_comparison} presents a comprehensive comparison between HTP and recent diffusion-based 3D HPE methods across multiple sampling configurations ($K{=}1$, $10$) with a fixed inference horizon ($H{=}20$). Under the low-sampling regime ($K{=}1$), HTP achieves a substantial acceleration of 3$\times$ in inference FPS while reducing per-frame MACs by over 56\%, along with the lowest MPJPE of 32.9. As $K$ increases, HTP consistently achieves both lower error and lower inference cost compared to all prior methods, demonstrating its robustness to different sampling budgets. Notably, even with $K{=}1$, our method surpasses the accuracy of several baselines operating at $K{=}10$ (\eg, 32.9 MPJPE vs. 35.4 of D3DP), while reducing MACs from 457.6G to 20.0G and boosting FPS from 142.2 to 2443.9—representing a 16$\times$ speedup. HTP thus achieves strong generalization and cost-efficiency across budget settings, enabling real-world deployment.

\begin{table}[t]
    \centering
    \caption{Efficiency Comparison with Non-Diffusion Baselines on Human3.6M.}
    \renewcommand{\arraystretch}{1.2}
    \resizebox{1.0\linewidth}{!}
    {
    \begin{tabular}{l c|ccc}
        \toprule
        \textbf{Method} & Type & MPJPE$\downarrow$ & Params (M) & MACs/frame \\
        \midrule
        PoseFormer~\cite{poseformer} & Transformer & 44.3  & 9.5 & 1.62 G \\
        PoseFormerV2~\cite{poseformerv2} & Transformer & 45.2  & 9.5 & 2.10 G \\
        STCFormer~\cite{stcformer}  & Transformer   & 40.8  & 18.9  & 0.32 G \\
        \rowcolor{black!10} HTP w/ MixSTE & Transformer   & 39.9  & 36.4  & 0.36 G \\
        \midrule
        \rowcolor{gray!10} \multicolumn{5}{l}{\textbf{\textit{Setting:} $K = 1$, $H = 1$}} \\
        D3DP~\cite{d3dp}   & Diffusion   & 40.0  & 34.6  & 1.14 G \\
        \rowcolor{black!10} HTP (ours) w/ $n_13$ & Diffusion & 39.8  &  37.5  & 0.72 G \\
        \rowcolor{black!10} HTP (ours) w/ $n_1=1$ & Diffusion & 40.6  &  37.5  & 0.50 G \\
        \bottomrule
    \end{tabular}
    }
    \label{tab:comparison_w_transformer}
\end{table}

\textbf{\textit{4) Efficiency Comparison with Non-Diffusion Baselines}:} We compare our diffusion-based HTP with representative Transformers in Tab.~\ref{tab:comparison_w_transformer}. Compared to the seq-to-frame baseline PoseFormer~\cite{poseformer} (1.62 G/frame), our HTP ($n_1=1$) is drastically more efficient (0.50 G/frame) and accurate (40.6 mm vs. 44.3 mm). Against the efficient seq-to-seq STCFormer~\cite{stcformer} (0.32 G/frame), our method achieves higher accuracy (40.6 mm vs. 40.8 mm) at a comparable computational scale. Moreover, our high-fidelity setting ($n_1=3$) further reduces the error to 39.8 mm while maintaining an affordable cost (0.72 G/frame), effectively positioning diffusion models as a high-performance competitor to lightweight transformers.

\begin{table}[t]
    \centering
    \begin{minipage}[t]{0.48\linewidth} 
        \centering
        \caption{Ablation Study on the Location of MGPTP Module on the Human3.6M Dataset.}
        \renewcommand{\arraystretch}{1.2}
        \setlength{\tabcolsep}{2pt} 
        \begin{tabular}{l c|c c}
            \toprule
            $f$ & $n_{1}$  & MPJPE $\downarrow$ & MACs (G) \\
            \midrule
            54 & 1 & 33.0 & 60.6 \\
            54 & 2 & 32.0 & 74.2 \\
            54 & 3 & 29.9 & 87.7 \\
            54 & 4 & 30.8 & 101.2 \\
            \bottomrule
        \end{tabular}
        \label{tab:ablation1}
    \end{minipage}%
    \hfill 
    \begin{minipage}[t]{0.48\linewidth}
        \centering
        \caption{Effect of the Sequence Length After MGPTP Module on the Human3.6M Dataset.}
        \renewcommand{\arraystretch}{1.2}
        \setlength{\tabcolsep}{2pt} 
        \begin{tabular}{l c|c c}
            \toprule
            $f$ & $n_{1}$  & MPJPE $\downarrow$ & MACs (G) \\
            \midrule
            27 & 3 & 31.2 & 79.7 \\
            54 & 3 & 29.9 & 87.7 \\
            81 & 3 & 30.1 & 95.6 \\
            108 & 3 & 32.1 & 103.6 \\
            \bottomrule
        \end{tabular}
        \label{tab:ablation2}
    \end{minipage}
\end{table}

\begin{figure}[t]
\centering
\includegraphics[width=0.9\columnwidth]{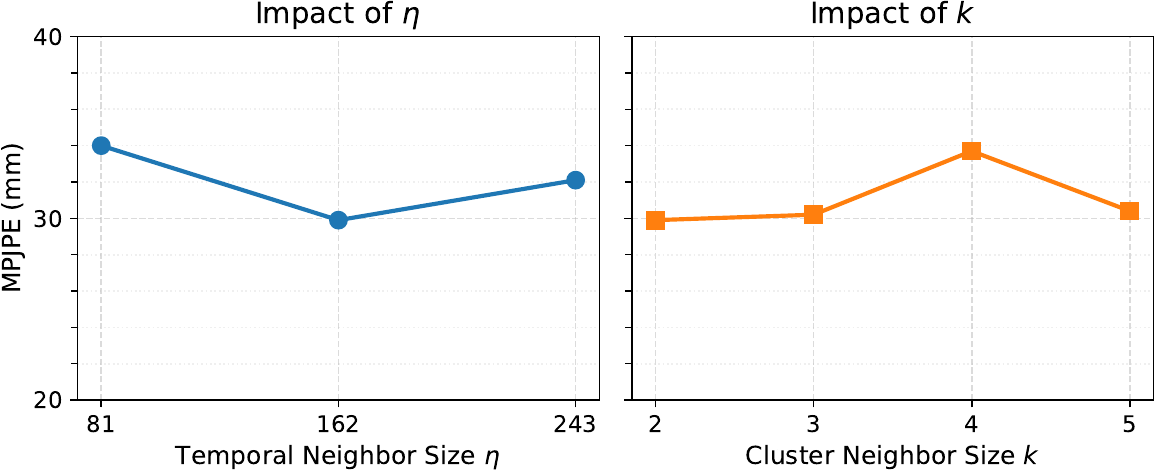}
\caption{The sensitivity analysis of $\eta$ and $k$.}
\label{fig_param_eta_k}
\end{figure}

\subsection{Sensitivity Analysis and Ablation Study}
We conduct comprehensive sensitivity and ablation studies on the Human3.6M dataset to validate the design of HTP. Specifically, we examine: (1)--(3) key architectural hyperparameters, including the pruned sequence length $f$, module placement $n_1$, and neighborhood sizes ($\eta, k$); (4) the flexibility of adjusting parameters ($\eta, n_1$) at inference time to balance efficiency; (5)--(6) the incremental contribution of each module and the specific role of the sparse mask $\mathbf{M}$ in guiding attention and pruning; and (7) the impact of input sequence length $F$ on the trade-off between performance and computational cost.

\textbf{\textit{1) Sequence length after MGPTP Module}:} The number of representative pose tokens retained by the MGPTP module plays a crucial role in performance. As shown in Tab. \ref{tab:ablation1}, increasing the sequence length does not lead to the anticipated performance improvements. This is because retaining too many pose tokens during clustering can amplify the impact of discrete or less informative tokens, which disrupts the coherent understanding of overall motion patterns. Thus, we set $f=54$ in all experiments to achieve an optimal balance.

\textbf{\textit{2) Location of MGPTP}:} In Tab. \ref{tab:ablation2}, we explore different settings for the number of SFT MHSA layers $n_{1}$, which effectively determines the placement of the MGPTP module. Adjusting $n_{1}$ allows for a trade-off between computational efficiency and performance. From the results, we observe that lower values of $n_{1}$ lead to reduced performance, as pruning pose tokens too early limits the network’s ability to learn aggregated motion information through SFT MHSA. Conversely, setting $n_{1}$ too high can result in attention collapse, which negatively impacts performance. We find that $n_{1}=3$ provides the best balance between efficiency and accuracy, and thus use this setting for all experiments.

\textbf{\textit{3) Hyperparameters $\eta$ and $k$}:} As visualized in Fig.~\ref{fig_param_eta_k}, we analyze the sensitivity of $\eta$ (in TCEP) and $k$ (in MGPTP). For $\eta$, performance peaks at $\eta=162$; lower values (\eg, 81) fail to capture sufficient context, while full sequences ($\eta=243$) introduce redundancy that negates pruning benefits. Regarding $k$, a compact neighborhood ($k=2$) yields the lowest MPJPE (29.9 mm). Increasing $k$ tends to over-smooth local density estimates, obscuring distinct motion states. Consequently, we adopt $\eta=162$ and $k=2$ as the default configuration.

\textbf{\textit{4) Inference Hyperparameters}:} We further evaluate the impact of adjusting hyperparameters during the inference phase. As shown in the Tab. \ref{tab:addition_ablation2}, for the temporal node number $\eta$, the optimal performance is achieved when the inference value matches the training setting ($\eta=162$). Regarding the number of SFT MHSA layers $n_1$ (Tab. \ref{tab:addition_ablation3}), although matching the training depth ($n_1=3$) yields the marginal best MPJPE, it incurs higher computational costs. We thus adopt $n_1=1$ as the default inference configuration for efficiency, achieving a significant FPS boost and MACs reduction with only a negligible performance trade-off compared to $n_1=3$.

\begin{table}[t]
    \centering
    \begin{minipage}[t]{0.4\linewidth}
        \centering
        \caption{Adjusting the Temporal Node Number $\eta$ in TCEP During Inference.}
        \renewcommand{\arraystretch}{1.2}
        \setlength{\tabcolsep}{3pt} 
        \begin{tabular}{c|c c}
            \toprule
            $\eta$ & MPJPE & FPS \\
            \midrule
            243 & 30.3 & 242.5 \\
            162 & 29.9 & 244.6 \\
            81 & 34.4 & 254.3 \\
            \bottomrule
        \end{tabular}
        \label{tab:addition_ablation2}
    \end{minipage}%
    \hspace{0.01\linewidth}
    \begin{minipage}[t]{0.4\linewidth}
        \centering
        \caption{Adjusting the Number of SFT MHSA Layers $n_{1}$ During Inference.}
        \renewcommand{\arraystretch}{1.2}
        \setlength{\tabcolsep}{2pt} 
        \begin{tabular}{c|c c c}
            \toprule
            $n_1$ & MPJPE & MACs & FPS \\
            \midrule
            1 & 29.9 & 99.8 & 244.6 \\
            2 & 29.7 & 122.0 & 204.5 \\
            3 & 29.1 & 228.8 & 176.5 \\
            \bottomrule
        \end{tabular}
        \label{tab:addition_ablation3}
    \end{minipage}%
\end{table}

\begin{table}[t]
    \centering
    \caption{Ablation Study on Different Designs of HTP.}
    \renewcommand{\arraystretch}{1.2}
    \setlength{\tabcolsep}{2pt} 
    \begin{tabular}{l |ccc| ccc}
        \toprule
        \multirow{2}*{Setting} & \multirow{2}*{TCEP} & SFT  & \multirow{2}*{MGPTP} & \multirow{2}*{MPJPE$\downarrow$} & \multirow{2}*{Param} & \multirow{2}*{MACs} \\
        & &  MHSA&  &  \\
        \midrule
        Baseline &  &  &  & 34.7 & 35.5 & 143.375 \\
        Setting1& \checkmark &  &  & 31.9  & 36.3 \textcolor{gray}{$\uparrow_{0.8}$} & 143.392 \textcolor{gray}{$\uparrow_{0.017}$} \\
        Setting2& \checkmark & \checkmark &  & 33.3 & 36.3 \textcolor{gray}{$\uparrow_{0.0}$} & 143.392 \textcolor{gray}{$\uparrow_{0.000}$} \\
        Setting3& \checkmark &  & \checkmark & 31.6 & 37.5 \textcolor{gray}{$\uparrow_{2.0}$} &87.648 \textcolor{gray}{$\downarrow_{55.727}$} \\
        \midrule
        \rowcolor{black!10} HTP (Ours) & \checkmark & \checkmark & \checkmark & 29.9 & 37.5 \textcolor{gray}{$\uparrow_{2.0}$} & 87.648 \textcolor{gray}{$\downarrow_{55.727}$} \\
        \bottomrule
    \end{tabular}   
    \label{tab:ablation3}
\end{table}

\begin{table}[t]
    \centering
    \caption{Analysis of the Impact of the Sparse Mask $\mathbf{M}$.}
    \renewcommand{\arraystretch}{1.2}
    \setlength{\tabcolsep}{4pt} 
    \begin{tabular}{l |cc|cc| c}
        \toprule
        \multirow{2}*{Setting} & \multicolumn{2}{c|}{Temporal MHSA}  & \multicolumn{2}{c|}{MGPTP} & \multirow{2}*{MPJPE$\downarrow$} \\
        \cmidrule(lr){2-5} 
        &  w $\mathbf{M}$ & w/o $\mathbf{M}$ & w $\mathbf{M}$ & w/o $\mathbf{M}$ &  \\
        \midrule
        Setting4 &  & \checkmark  &  &  \checkmark & 32.7   \\
        Setting5 &  & \checkmark  & \checkmark &  & 31.6 \\
        Setting6 & \checkmark  &  &  & \checkmark & 31.3 \\
        \midrule
        \rowcolor{black!10} HTP (Ours)  & \checkmark &  & \checkmark &  & 29.9  \\
        \bottomrule
    \end{tabular}   
    \label{tab:addition_ablation4}
\end{table}

\begin{table}[t]
    \centering
    \caption{Impact of Input Sequence Length $F$.}
    \renewcommand{\arraystretch}{1.2}
    \resizebox{0.85\linewidth}{!}
    {
    \begin{tabular}{l c|ccc}
        \toprule
        \multirow{2}{*}{F} & \multirow{2}{*}{Batch Size} &
        \multirow{2}{*}{MPJPE$\downarrow$} & 
        \multicolumn{2}{c}{Inference} \\
        \cmidrule(lr){4-5}
        & & & MACs/frame & FPS$\uparrow$ \\
        \midrule
        81 & 4 & 34.4 & 119.2 & 139.2 \\
        81 & 32 & 34.4 & 119.2 & \textbf{140.6} \\
        162 & 4 & 34.3 & 119.2 & 130.2 \\
        162 & 12 & 34.3 & 119.2 & 130.8 \\
        \midrule
        \rowcolor{black!10} 243 & 4 & \textbf{29.9} & \textbf{99.8} & 137.0 \\
        \bottomrule
    \end{tabular}
    }
    \label{tab:ablation_F}
\end{table}

\begin{figure*}[t]
    \centering
    \begin{minipage}{0.48\textwidth}
        \centering
        \includegraphics[width=\textwidth]{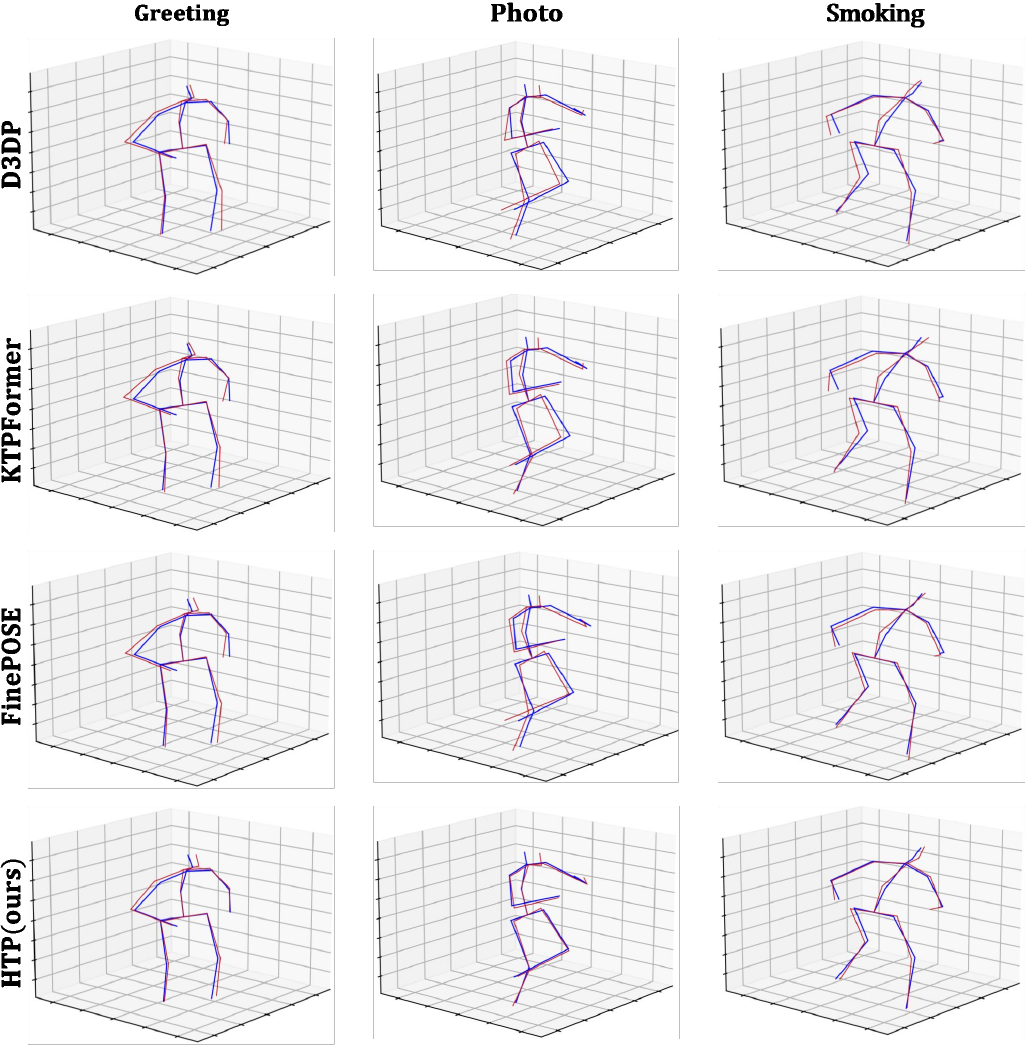}
        \caption{\textbf{Qualitative comparisons of our HTP with previous state-of-the-art methods \cite{d3dp, ktpformer, finepose} on the Human3.6M dataset.} Solid blue line: ground-truth 3D pose. Solid red line: estimated 3D pose.}
        \label{fig_vis}
    \end{minipage}
    \hfill
    \begin{minipage}{0.48\textwidth}
        \centering
        \includegraphics[width=\textwidth]{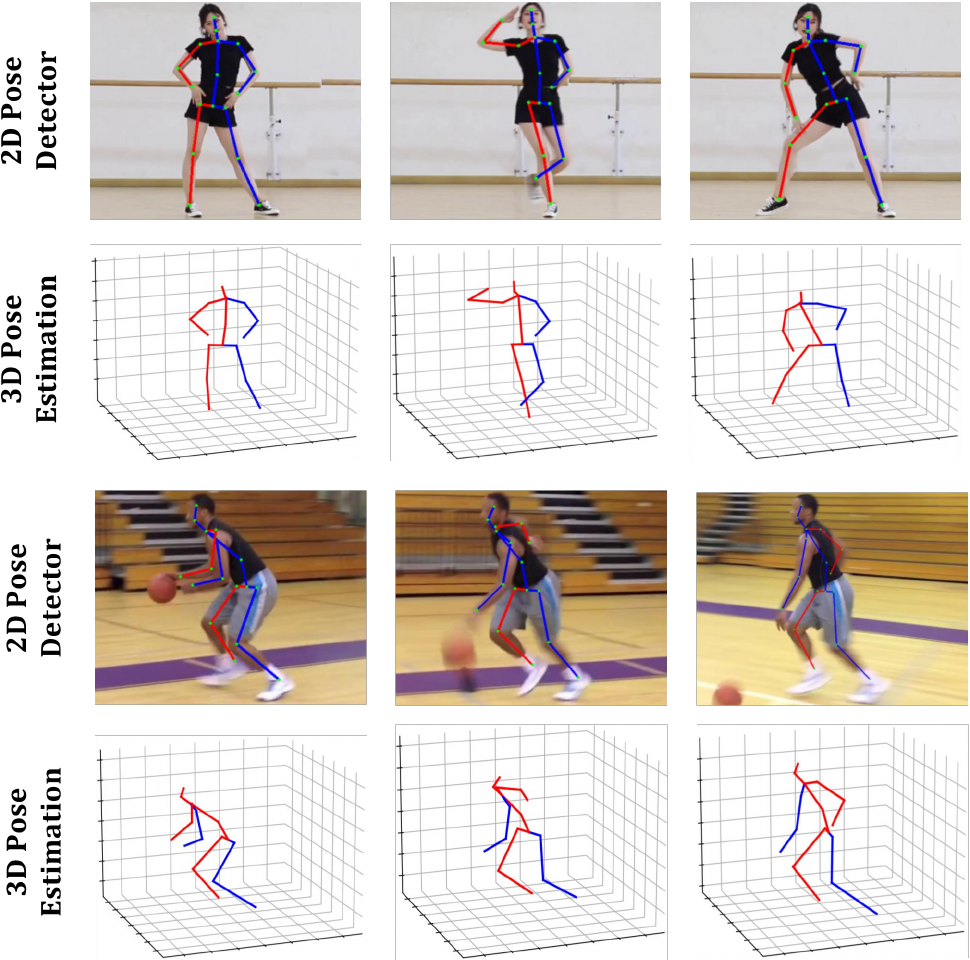}
        \caption{ \textbf{Qualitative results of the proposed method on in-the-wild videos.} Setting H=5, K=5. HRNet \cite{hrnet} was used as the 2D keypoint detector to generate 2D poses.}
        \label{in_the_wild}
    \end{minipage}
\end{figure*}

\begin{figure}[t]
\centering
\includegraphics[width=0.95\columnwidth]{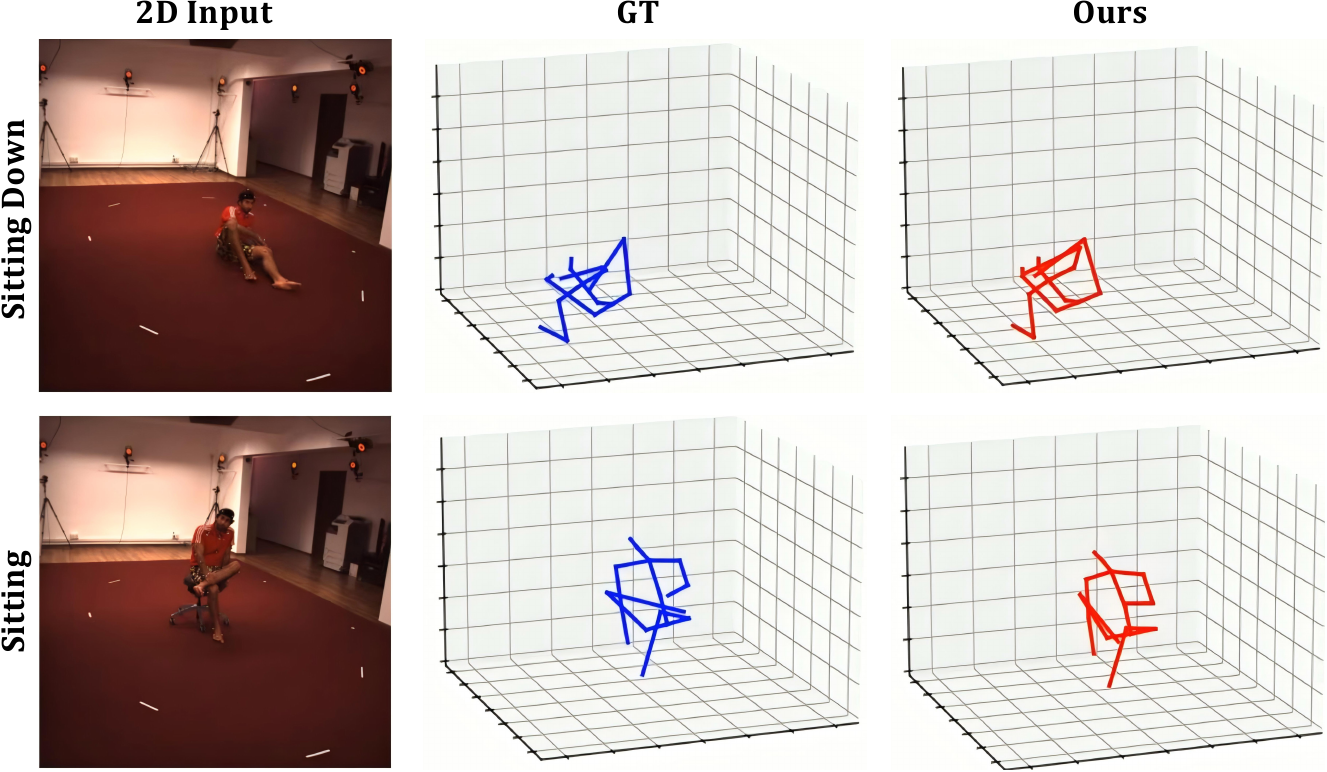}
\caption{Visualization of a failure case under severe self-occlusion.}
\label{fig_failure_case}
\end{figure}

\begin{figure*}[t]\centering
\includegraphics[width=0.9 \textwidth]{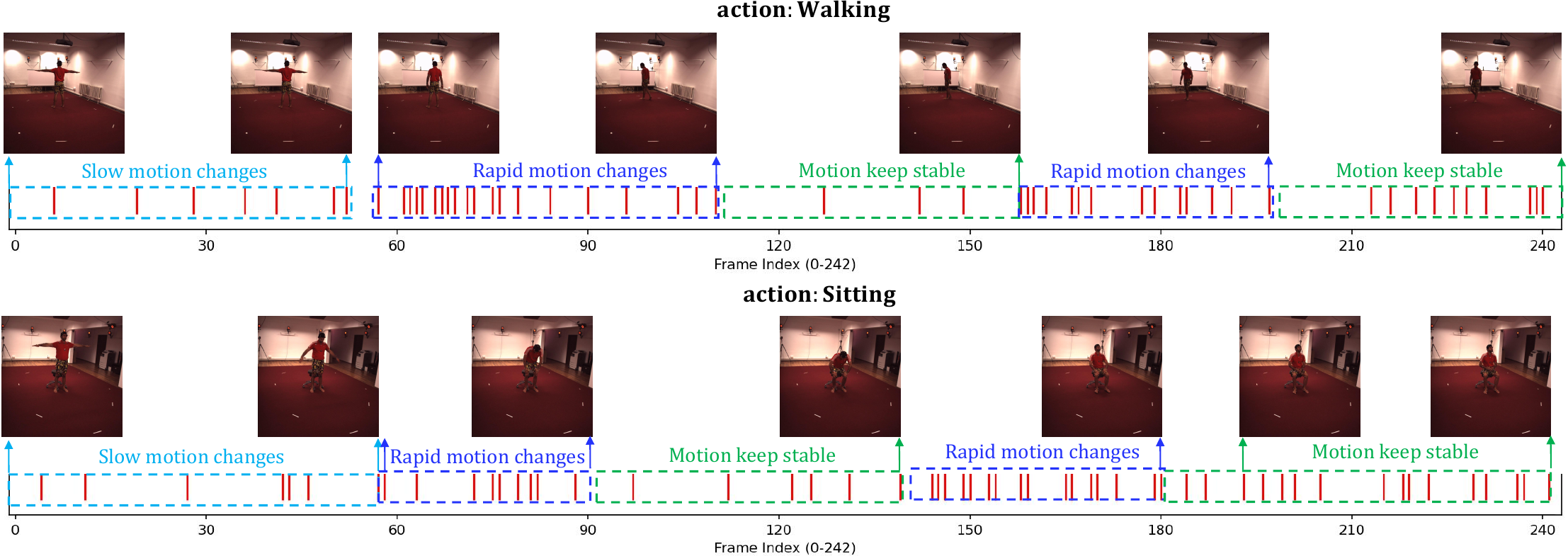}
\caption{\label{fig_token} \textbf{Qualitative Analysis of Frame Retention in HTP.} For each action, the timeline at the bottom shows the frame index (from 0 to 242), with red vertical lines marking the frames retained by HTP.}
\end{figure*}

\textbf{\textit{5) Effect of Each Module}:} 
Tab.~\ref{tab:ablation3} investigates the role of each module through incremental configurations. We begin with a baseline that extends D3DP \cite{d3dp} by adding the Spatial GCN from \cite{agformer}. In \textbf{Setting1}, we introduce the TCEP module without applying its generated sparse mask, resulting in a significant MPJPE reduction of 2.8 with only a minor increase in parameters (0.8M) and a marginal computational cost of 0.017 G MACs. This highlights TCEP’s effectiveness in capturing essential motion cues by dynamically selecting temporal dependencies. In \textbf{Setting2}, we further adopt the sparse mask from TCEP to replace full temporal attention with global SFT MHSA. However, this setting performs worse than Setting1, indicating that overly aggressive sparsification may lead to attention collapse and hinder temporal reasoning. These results suggest that \textit{moderate} application of sparse self-attention is essential to maintain global context. 

In \textbf{Setting3}, we remove the sparse mask from Setting2 and instead apply the MGPTP module to prune redundant tokens. By utilizing the sparsity mask generated by TCEP to compress the sequence length, this setting achieves a net reduction of 55.7 G MACs, demonstrating that the efficiency gains from hierarchical pruning vastly outweigh the minimal overhead of the pruning decision modules. Finally, the full HTP configuration combines selective sparse attention with MGPTP, achieving the optimal balance between reconstruction accuracy and computational cost.

\textbf{\textit{6) Impact of the Sparse Mask $\mathbf{M}$}:} As shown in Tab.~\ref{tab:addition_ablation4}, we evaluate the role of the mask $\mathbf{M}$ generated by TCEP. Compared to the baseline without mask guidance (\textbf{Setting 4}), applying $\mathbf{M}$ solely to MGPTP (\textbf{Setting 5}) or SFT MHSA (\textbf{Setting 6}) yields moderate gains. However, the full HTP configuration, which leverages $\mathbf{M}$ in both modules, outperforms Setting 5 and Setting 6 by 1.7mm and 1.4mm, respectively. This demonstrates that sparse attention and mask-guided pruning are mutually reinforcing: selective attention enhances feature discriminability for pruning, while accurate token retention preserves global context. Thus, $\mathbf{M}$ serves as a critical structural bridge, integrating the modules into a cohesive pruning pipeline.

\textbf{\textit{7) Impact of Input Sequence Length $F$:}} We analyze the trade-off between sequence length and performance in Tab.~\ref{tab:ablation_F}. Following prior methods~\cite{mixste, d3dp,finepose,stcformer}, we adopt $F=243$ as the default to maximize long-range temporal context. Empirical results show that reducing $F$ to 162 or 81 degrades MPJPE. While shorter sequences reduce VRAM consumption and allow for larger batch sizes, our tests indicate that larger batch sizes yield minimal gains in inference FPS. Given that HTP's pruning strategy already ensures high efficiency for 243-frame inputs, we prioritize the accuracy benefits of the longer sequence, recommending shorter versions only for memory-constrained environments.

\subsection{Qualitative Analysis}
\textbf{\textit{1) Qualitative results comparison}:} Fig. \ref{fig_vis} compares HTP with state-of-the-art diffusion-based methods \cite{d3dp, ktpformer, finepose} on Human3.6M ($H=20, K=10$). Across actions ranging from simple poses (e.g., \textit{``Photo''}) to complex articulations (e.g., \textit{``Sitting down''}, \textit{``Smoking''}), HTP consistently yields 3D estimations that align closer to the ground truth. Notably, our method demonstrates superior fidelity in limb joints (e.g., elbows and wrists) and maintains better structural plausibility in challenging poses compared to baselines, effectively mitigating the joint distortions observed in D3DP and KTPFormer.

\textbf{\textit{2) In-the-wild Videos}:} To validate generalization, we evaluate HTP on wild videos using 2D poses detected by HRNet \cite{hrnet}. As shown in Fig. \ref{in_the_wild}, our method exhibits exceptional robustness, maintaining high accuracy even in challenging scenarios with severe self-occlusion and rapid motion.

\textbf{\textit{3) Qualitative Analysis of Frame Retention}:} Fig.~\ref{fig_token} visualizes the adaptive pruning behavior of HTP on \textit{``Walking''} and \textit{``Sitting''}. By selecting only 54 representative frames from the 243-frame input, HTP dynamically allocates computational resources based on motion complexity. Specifically, the model retains a higher density of frames during rapid transitions (blue dashed box) to capture fast-moving dynamics, while aggressively pruning frames during stable or slow-motion phases (cyan/green dashed boxes). This content-aware strategy effectively balances temporal sparsity with representational completeness,  maintaining accurate 3D pose estimation while significantly reducing computational overhead.

\section{Limitations and Future Works}
\label{sec:limation}
Despite HTP's substantial efficiency gains, specific challenges present avenues for improvement. First, as shown in Fig.~\ref{fig_failure_case}, severe self-occlusions can cause the pruning mechanism to inadvertently discard critical frames needed for resolving complex articulations. Second, as a 2D-to-3D lifting framework, HTP's performance is bounded by 2D input quality, limiting gains in noisy outdoor scenarios. Future work will address these by exploring occlusion-aware attention and spatial uncertainty modeling to enhance robustness.

\section{Conclusions}
\label{sec:conclusion}

In this paper, we address the efficiency challenges in diffusion-based 3D human pose estimation with Hierarchical Temporal Pruning (HTP). HTP progressively reduces redundancy by selectively pruning pose tokens across both frame and semantic levels while preserving critical motion dynamics. Starting with temporal correlations in TCEP, focusing attention on SFT MHSA, and refining through MGPTP’s semantic-level clustering, HTP selectively retains the most motion-critical pose tokens. Extensive experiments on Human3.6M and MPI-INF-3DHP datasets demonstrate that HTP achieves state-of-the-art accuracy while substantially reducing computational cost and improving inference speed.

\bibliographystyle{IEEEtran}
\bibliography{Manuscript}{}

\end{document}